\documentclass[sigconf, 10pt, screen, nonacm]{acmart}

\renewcommand{\ln}{\mathop{\mathrm{ln}}}
\renewcommand{\log}{\mathop{\mathrm{log}}}
\renewcommand{\exp}{\mathop{\mathrm{exp}}}

\usepackage{pifont}
\usepackage{enumitem}

\AtBeginDocument{
  }
\usepackage{threeparttable}
\usepackage{xspace}
\usepackage{subcaption}
\usepackage{graphicx}
\renewcommand\footnotetextcopyrightpermission[1]{}
\setcopyright{acmlicensed}
\copyrightyear{2018}
\acmYear{2018}
\acmDOI{XXXXXXX.XXXXXXX}
\acmISBN{}
\usepackage{adjustbox}

\usepackage{siunitx}

\newcommand{\cmark}{\textcolor{green}{\checkmark}}
\newcommand{\xmark}{\textcolor{red}{\ding{55}}}
\newcommand{\oursystem}{\textsc{VisRFTwin}\xspace}

\begin{document}

\title[Taming Vision Priors for Data Efficient mmWave Channel Modeling]{\LARGE Taming Vision Priors for Data Efficient mmWave Channel Modeling}

\author[Z. An, L. Shangguan, J. Kaewell, et al.]{\large
  Zhenlin An\textsuperscript{1,2},\quad
  Longfei Shangguan\textsuperscript{2},\quad
  John Kaewell\textsuperscript{3},\quad
  Philip Pietraski\textsuperscript{3}\\
  Jelena Senic\textsuperscript{4},\quad
  Camillo Gentile\textsuperscript{4},\quad
  Nada Golmie\textsuperscript{4},\quad
  Kyle Jamieson\textsuperscript{5}\\
  \textsuperscript{1}University of Georgia \enspace
  \textsuperscript{2}University of Pittsburgh \enspace
  \textsuperscript{3}Interdigital \enspace
   \textsuperscript{4}NIST \enspace \textsuperscript{5}Princeton University
}

\begin{abstract}
Accurately modeling millimeter-wave (mmWave) propagation is essential for real-time AR and autonomous systems. Differentiable ray tracing offers a physics-grounded solution but still facing deployment challenges due to its over-reliance on exhaustive channel measurements or brittle, hand-tuned scene models for material properties. We present \oursystem, a scalable and data-efficient digital-twin framework that integrates vision-derived material priors with differentiable ray tracing. Multi-view images from commodity cameras are processed by a frozen Vision-Language Model to extract dense semantic embeddings, which are translated into initial estimates of permittivity and conductivity for scene surfaces. These priors initialize a Sionna-based differentiable ray tracer, which rapidly calibrates material parameters via gradient descent with only a few dozen sparse channel soundings. Once calibrated, the association between vision features and material parameters is retained, enabling fast transfer to new scenarios without repeated calibration. Evaluations across three real-world scenarios, including office interiors, urban canyons, and dynamic public spaces show that \oursystem reduces channel measurement needs by up to 10$\times$ while achieving a 59\% lower median delay spread error than pure data-driven deep learning methods.
\end{abstract}

\maketitle
{\renewcommand{\thefootnote}{}\footnotetext{Certain commercial equipment, instruments, or materials are identified in this paper in order to specify the experimental procedure adequately. Such identification is not intended to imply recommendation or endorsement by National Institute of Standards and Technology (NIST), nor is it intended to imply that the materials or equipment identified are necessarily the best available for the purpose.}}
\vspace{-2mm}
\section{Introduction}
\label{s:intro}

Millimeter-wave (mmWave) communication is a cornerstone of NextG wireless systems, promising multi-gigabit throughput and sub-millisecond latency~\cite{jain_two_2021, woodford_spacebeam_2021}. These capabilities are critical for emerging applications such as AR/VR~\cite{woodford_spacebeam_2021}, autonomous driving~\cite{kamari_mmsv_2023}, and dense IoT deployments~\cite{jain_two_2021}. Yet mmWave links remain fragile due to extreme free-space loss, sparse multipath, and sensitivity to blockage and mobility~\cite{mi_measurement-based_2024, narayanan_lumos5g_2020}. Reliable operation thus requires accurate and scalable channel modeling to manage directional beams and seamless handovers~\cite{woodford_spacebeam_2021,wang_demystifying_2020}.

Deep learning–based channel modeling solutions treat the radio environment as a black box, directly learning the mapping from transceiver locations to channel responses~\cite{mi_measurement-based_2024}. While these methods show promise in small, controlled testbeds, they demand thousands of site-specific channel measurements for model training (e.g., 2000 samples for a $200\text{m}^2$ room~\cite{mi_measurement-based_2024}, or 30k for a single city block in Lumos5G~\cite{narayanan_lumos5g_2020}), making them impractical at scale. Moreover, models such as \textsc{PointNet}-based solution~\cite{mi_measurement-based_2024} or the sequence-to-sequence regressors in Lumos5G~\cite{narayanan_lumos5g_2020} generalize poorly when the environment changes, forcing retraining in every new deployment.
\begin{figure}[t!]
    \centering
    \includegraphics[width=0.95\linewidth]{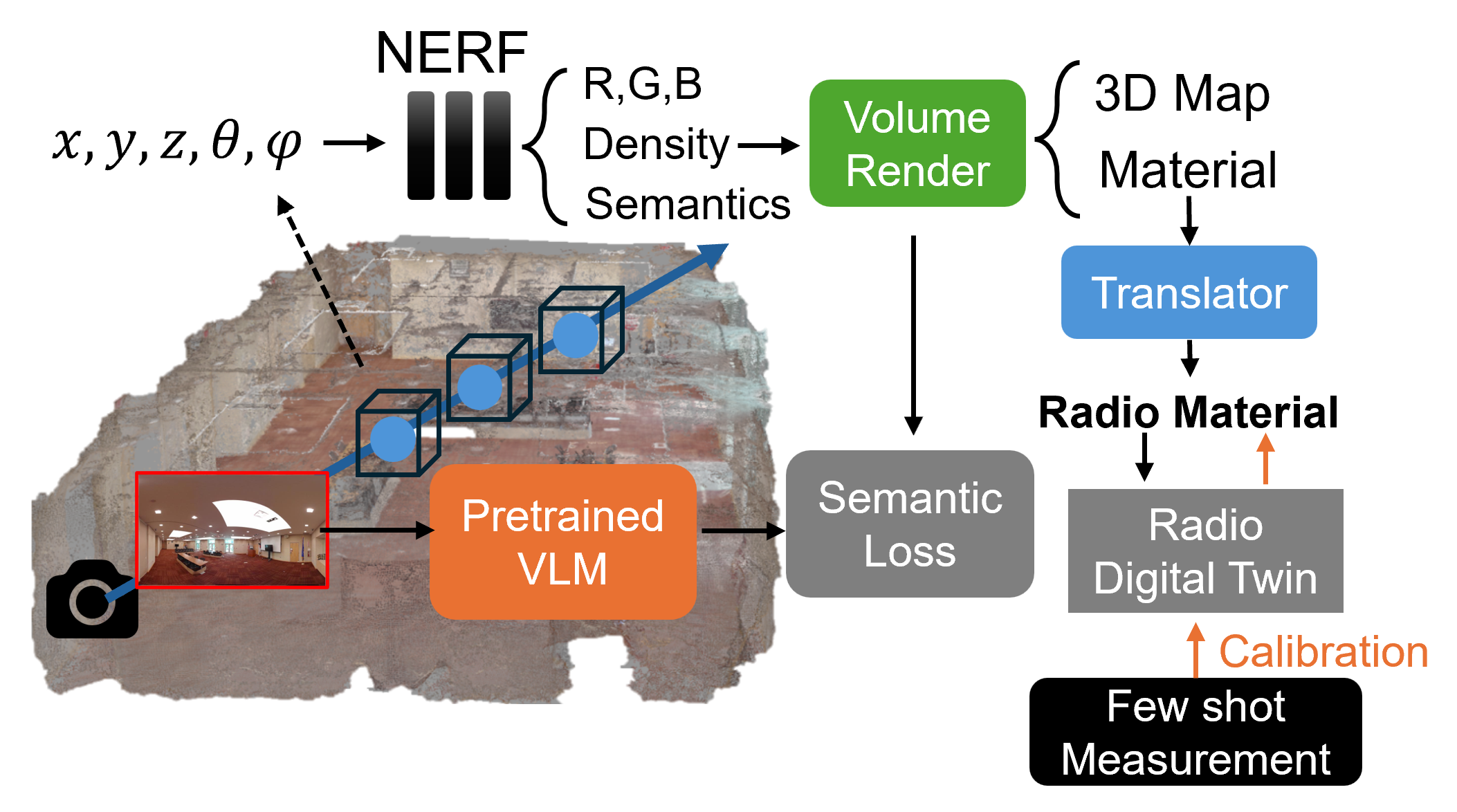}\vspace{-4mm}
    \caption{System Workflow. \textnormal{\oursystem leverages 3D scene reconstruction (NeRF) and pretrained vision–language models to extract semantic material features from images. These vision priors are translated into electromagnetic parameters to bootstrap differentiable ray tracing,  reducing the amount of channel measurements needed for channel modeling.}}
    \vspace{-6mm}
    \label{fig:sys-overview}
\end{figure}

Recent years have witnessed the surge of physics-inspired deep learning for wireless channel modeling~\cite{zhao_nerf2_2023, orekondy_winert_2022, chen_rfcanvas_2024}. These approaches improves accuracy, interpretability, and data efficiency of channel modeling by embedding the electromagnetic (EM) parameter priors into neural networks. Broadly, they fall into two categories (Table~\ref{tab:rf_sim_compare}): (i) optical ray-marching methods, such as NeRF2~\cite{zhao_nerf2_2023}, which approximate radio propagation using optical volumetric rendering; and (ii) differentiable ray tracing, which explicitly models physical RF signal propagation paths and backpropagates to learn the EM parameters, such as  such as the conductivity and permittivity, of each individual objects in the environment~\cite{an_radiotwin_2025, hoydis_learning_2024}.

However, optical ray-marching methods fail to accurately characterizing RF signal propagation. They also cannot recover multipath delay profiles, which limits their practical use in wideband wireless communication and sensing systems. While the differentiable ray tracing methods can accurately model the RF signal propagation, they require substantial amount of channel measurements (\S\ref{sec:background}) to calibrate the EM parameters of the entire environment. Moreover, these methods cannot deal with environmental dynamics.

\vspace{2mm}
\noindent
In this paper, we aim to retain the high physical fidelity of differentiable ray tracing method while eliminating its over-reliance on large amount channel measurements for calibration. Taking a step further, our goal is to empower differentiable ray tracing to accurately model multipath RF channel profile with only a handful of measurements. By drastically reducing the need for extensive site-specific channel measurements, we believe our work would significantly improve the scalability of differentiable ray tracing, making it a major step toward practical deployment.

Our work is motivated by two insights, one regarding a fundamental physical law and the other a notable research trend. (i) the ray–object surface interactions (e.g., reflection) are governed by an object’s electromagnetic parameters (e.g., conductivity, permittivity) which in turn are determined by its physical attributes such as texture and material composition (e.g., wood, concrete)\footnote{\textit{Surface-level} reflection and diffuse scattering dominate mmWave propagation according to prior works~\cite{mi_measurement-based_2024, wang_demystifying_2020, woodford_spacebeam_2021}. Thus, visual surface texture is the primary cue needed for estimating the material's EM properties.} and (ii) recent advances in computer vision enable open-vocabulary recognition of textures and materials via large vision–language models~\cite{wu_how_2022,openai_gpt4_2023}.
This leads to the question: {\it can we exploit vision-derived priors to initialize electromagnetic coefficients of environmental objects in the differentiable ray-tracing framework, enabling rapid calibration with substantially fewer channel measurements?}
\begin{table}[t!]
\centering
\caption{Comparison of channel modeling methods.}\vspace{-4mm}
\begin{minipage}{\linewidth}
\centering
\resizebox{\linewidth}{!}{
\begin{tabular}{l|c|c|c|c|c|c|c}
\toprule
\textbf{Approach} &
\textbf{Model} &
\shortstack{\textbf{Physics}\\\textbf{Fidelity}} &
\shortstack{\textbf{RF Data}\\\textbf{Needed}} &
\shortstack{\textbf{Vision}\\\textbf{Modality}}  &
\shortstack{\textbf{Scene}\\\textbf{Dynamics}}&
\shortstack{\textbf{Multipath}\\\textbf{Delay}} &
\shortstack{\textbf{Scattering}\\\textbf{Physics}} \\
\midrule
PointNet~\cite{mi_measurement-based_2024} &
e2e DL &
Low &
Dense &
\xmark &
\xmark &
\cmark &
\xmark \\
Sionna~\cite{hoydis_sionna_2023} &
DiffRT (RF Prop.)&
High &
Medium &
\xmark &
\xmark &
\cmark &
\cmark \\

RadioTwin~\cite{an_radiotwin_2025} &
DiffRT (RF Prop.)&
High &
Dense &
\xmark &
\xmark &
\cmark &
\cmark \\
\hline
NeRF2~\cite{zhao_nerf2_2023} &
RM (Optical Prop.) &
Medium &
Dencs &
\xmark &
\xmark &
\xmark &
\xmark \\

RF-3DGS~\cite{zhang_rf-3dgs_2025} &
RM (Optical Prop.) &
Medium &
Medium &
\cmark &
\xmark &
\xmark &
\xmark \\

RFCanvas~\cite{chen_rfcanvas_2024} &
RM (Optical Prop.) &
Medium &
Sparse &
\cmark &
\cmark &
\xmark &
\xmark \\
\midrule

\textbf{Ours} &
\textbf{DiffRT (RF Prop.)} &
\textbf{High} &
\textbf{Sparse} &
\textbf{\cmark} &
\textbf{\cmark} &
\textbf{\cmark} &
\textbf{\cmark} \\
\bottomrule
\end{tabular}
}
\scriptsize{\textbf{Note:}
DL: Deep Learning,
DiffRT: Differentiable Ray Tracing,
RM: Ray Marching
}\vspace{-5mm}
\end{minipage}
\label{tab:rf_sim_compare}
\end{table}

We give an affirmative answer by presenting \oursystem: a data-efficient, vision-guided digital-twin framework.
\oursystem works by
identifying each environmental object's fine-grained material properties in the 3D space, and correlating these material properties with their EM parameters based on wave-propagation physics.

As sketched in Fig.\ref{fig:sys-overview}, \oursystem leverages multi-view RGB images to train a \emph{semantic NeRF}, which predicts not only color and density, but also a high-dimensional feature grid whose voxel embeddings are forced to match CLIP (Contrastive Language-Image Pre-Training) features extracted from every input view via a semantic loss.  The resulting 3-D volume simultaneously provides centimeter-accurate geometry and open-vocabulary material cues.  A lightweight translator is then proposed to convert each voxel’s learned features into continuous, frequency-dependent electromagnetic parameters that seed a differentiable Sionna ray tracer~\cite{hoydis_sionna_2023} modelling reflection, scattering, and diffraction.  With only a few dozen sparse channel soundings, gradient back propagation jointly refines geometry and material parameters, producing an environment-specific twin without exhaustive retraining. To achieve this goal, there are multiple challenges:

\noindent$\bullet$ First, multi-view 2D RGB views must be ``lifted'' into a 3-D semantic map that retains both good geometry and rich semantic features that govern radio object interactions.
Existing pixel-level feature extractors, however, often blur information across views and discard fine-grained semantics.
We address this by pairing a frozen vision–language backbone (CLIP) with a NeRF-based reconstruction loop and a pyramid feature-extraction strategy that processes multi-scale batches to compute per-pixel embeddings, enforcing view consistency while distilling radio-relevant semantics.

\noindent$\bullet$ Second, existing VLM based segmentation algorithms like Segment3D~\cite{cen_segment_2024,huang2024segment3d} can only extract an object's semantic label, such as wooden table, leather sofa, but not their EM parameters such as permittivity, conductivity, and surface-roughness. Hence they cannot be directly leveraged to model the ray-object interaction for RF channel modeling and prediction. To address this, prior works~\cite{kanhere_calibration_2023} hard-code a set of EM parameters for a given scene yet lack generalizability. We introduce a lightweight translator: a physics-regularized latent material prior that maps open-vocabulary object labels to frequency-dependent EM parameters.

\noindent$\bullet$ Third, our ambient environment changes over time, as real spaces evolve (e.g., people walking), yet both classical and differentiable ray tracers typically assume static meshes. Our twin maintains fidelity by incrementally updating the 3D map and re-optimizing local radio material. To further handle such dynamics, we localize changes to the affected regions and perform targeted refinements rather than re-optimizing the entire scene, ensuring both efficiency and adaptability.

We evaluate \oursystem using a set of real-world dataset we collected from (1) a 20~m $\times$ 10~m lecture room~\cite{mi_measurement-based_2024} (4956 samples, LoS-dominated) furnished with tables and chairs; (2) a 20~m $\times$ 15~m lobby (5100 samples, including non-LoS paths) containing sofas, tables, walls, and glass panels~\cite{gentile_context-aware_2024}; and (3) an outdoor scenario illustrated in Figure~\ref{fig:outdoor-ray}. The experimental results show that \oursystem\ reduces channel measurement requirements by up to $10\times$ while maintaining comparable accuracy, and achieves a 59\% lower median delay spread error in the zero-shot setting compared to data-driven models trained on 20\% of the dataset; maintains stable performance across LoS and NLoS regions; and, across office interiors, urban canyons, and dynamic public spaces.

\noindent \textbf{Contribution.} We make the following contributions.
\begin{itemize}[leftmargin=10pt, topsep=1pt]
    \item We propose a novel cross-modality digital-twin framework that explores the vision primitives to derive the environmental objects' electromagnetic properties.
    \item We integrate this vision priors into Sionna differentiable ray tracing framework, enabling rapid refinement with substantially fewer channel measurements.
    \item We evaluate the system across two indoor environment, one urban street, and one dynamic public space, and demonstrate the proposed system requires 10$\times$ less channel measurements while achieving significantly higher multipath profile modelling error than baselines.
\end{itemize}

Note that we are not the first work exploring visual cues for accurate channel modeling.
Vision-guided ray-marching methods such as RF-3DGS~\cite{zhang_rf-3dgs_2025} and RFCanvas~\cite{chen_rfcanvas_2024} have already leveraged multi-view RGB images to impose geometric or material reflection coefficient on RF signal propagation. These methods reconstruct a coarse 3-D scene representation through 3D vision techniques, then \textit{march rays using an optical rendering surrogate}. Although this substantially reduces the need for exhaustive RF measurements and yields strong environmental priors, the surrogate remains governed by optical, not electromagnetic physics. Consequently, it cannot accurately model material-dependent RF signal reflections, diffraction, and scattering phenomenon, nor can it produce physically meaningful multipath delays. Thus, while vision-prior ray–marching improves data efficiency, it is fundamentally incapable of recovering the full EM-consistent multipath required for accurate channel prediction. Table~\ref{tab:rf_sim_compare} summarizes the difference of \oursystem and these systems.

\begin{figure}[t]
    \centering
    \begin{subfigure}[t]{0.75\linewidth}
        \centering
        \includegraphics[width=\linewidth]{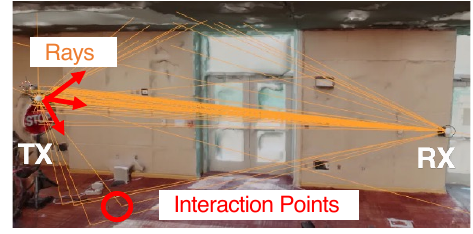}\vspace{-3mm}
        \caption{Ray tracing process.}
        \label{fig:diff-ray-loop}
    \end{subfigure}
    \begin{subfigure}[t]{\linewidth}
        \centering
        \includegraphics[width=\linewidth]{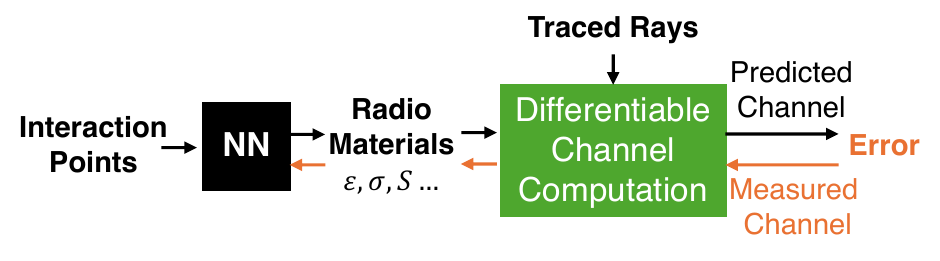}\vspace{-3mm}
        \caption{Differentiable ray tracing loop.}
        \label{fig:diff-ray-process}
    \end{subfigure}
    \vspace{-5mm}
    \caption{Differentiable ray tracing framework.
    \textnormal{(a) Real-world ray tracing visualization with rays, transmitters, receivers, and interaction points. (b) End-to-end differentiable training loop using channel mismatch as gradient feedback.
    }}
    \vspace{-4mm}
    \label{fig:diff-ray}
\end{figure}

\vspace{-3mm}
\section{Primer: Differentiable Ray Tracing}
\label{sec:background}

{\it Differentiable ray tracing} is designed for calibrating wireless channel simulations to match physical reality~\cite{zhao_nerf2_2023, hoydis_sionna_2023, orekondy_winert_2022, lu_newrf_2024}. The core idea is to construct the entire simulation pipeline (i.e., from scene geometry and material properties to wave propagation physics) in a way that is fully differentiable.

As shown in Fig.~\ref{fig:diff-ray}, given the 3D mesh representation of a scene, the ray tracer shoots massive numbers of rays to find those signal paths to the receiver.
The framework then uses a neural network called radio material neural network to map each ray-object interaction point to its specific EM parameters, like the conductivity $\sigma$, permittivity $\epsilon_r$ of the wall reflector.
This information, along with the identified paths, is used to compute the wireless channel \cite{an_radiotwin_2025}. To train the aforementioned neural network, a loss function is designed to  quantify the error between this computed channel and the actual channel measurement. During training, the framework uses a large number of channel measurements to refine the neural network's parameters via gradient descent.

Unlike the "black box" approach of DNNs, differentiable ray tracing leverages fundamental RF propagation principles like reflection and scattering, which allows it to generalize better to new environments~\cite{zhao_nerf2_2023, orekondy_winert_2022}. The optimized parameters, such as material's permittivity and conductivity, are also physically meaningful, providing higher interpretability and a deeper understanding of the environment, a key benefit that end-to-end deep learning models lack.

\noindent\textbf{The Limitations of Differentiable Ray Tracing}. Despite its interpretability, differentiable ray tracing framework remains heavily data-dependent. Since the EM properties of each individual environmental object are unknown a priori, the framework ought to learn them from scratch through gradient updates driven by significant amount of channel measurements. In our preliminary tests on the NIST's lobby dataset, we observed that the channel estimation errors exceeded 8~dB when using fewer than 50 channel measurements for calibration. This error converges to around 2~dB after several hundred measurements. Thus, while differentiable ray tracing scales better than end-to-end learning, its over-reliance on large site-specific channel measurement campaigns limits practicality in real deployments.

\vspace{-3mm}
\section{System Overview}
\label{s:system_overview}
Rather than learning EM parameters from scratch, we propose using each object's visual cues to provide a reasonable initial 'guess'.
We believe this is feasible because the EM parameters are fundamentally determined by an object's physical attributes, like texture and material composition (e.g., wood, concrete), which are inherently distinguishable through visual features.
These initial ``guessed'' EM parameters are used to initialize the radio parameter neural network in the differentiable ray tracing framework and then refined gradually using significantly less amount of channel measurements, thereby boosting the material neural network's training process while retaining physical interpretability.
As shown in Fig.~\ref{fig:sys-overview}, \oursystem consists of three stages:

\noindent(1) \textbf{Semantic Feature Distillation}. We begin by reconstructing a 3D scene from multi-view images (e.g., through NeRF) and sampling points on every environmental surface. Since there is no pre-trained vision models that can directly output the EM parameters of each object based on its visual input, we thus leverage pre-trained vision-language models (e.g.\ CLIP) to first extract the visual-textual semantic embeddings that capture each object's categories, textures, and spatial context on point-level granularity, without the need of any environment-dependent tuning.

\noindent(2)~\textbf{Visual-to-EM Parameter Translator.} We then design a lightweight translator that maps the per-point visual semantic embeddings into EM properties grounded in empirical models. It outputs the relative permittivity, conductivity, and scattering coefficients, which are parameterized following canonical electromagnetic relations (e.g., roughness-dependent scattering loss models~\cite{mukherjee_scalable_2022}). By constraining the translator with these physics-informed equations, \oursystem ensures that vision-derived material cues are converted into radio parameters consistent with established EM theory.

\begin{figure}[t]
    \centering
    \includegraphics[width=\linewidth]{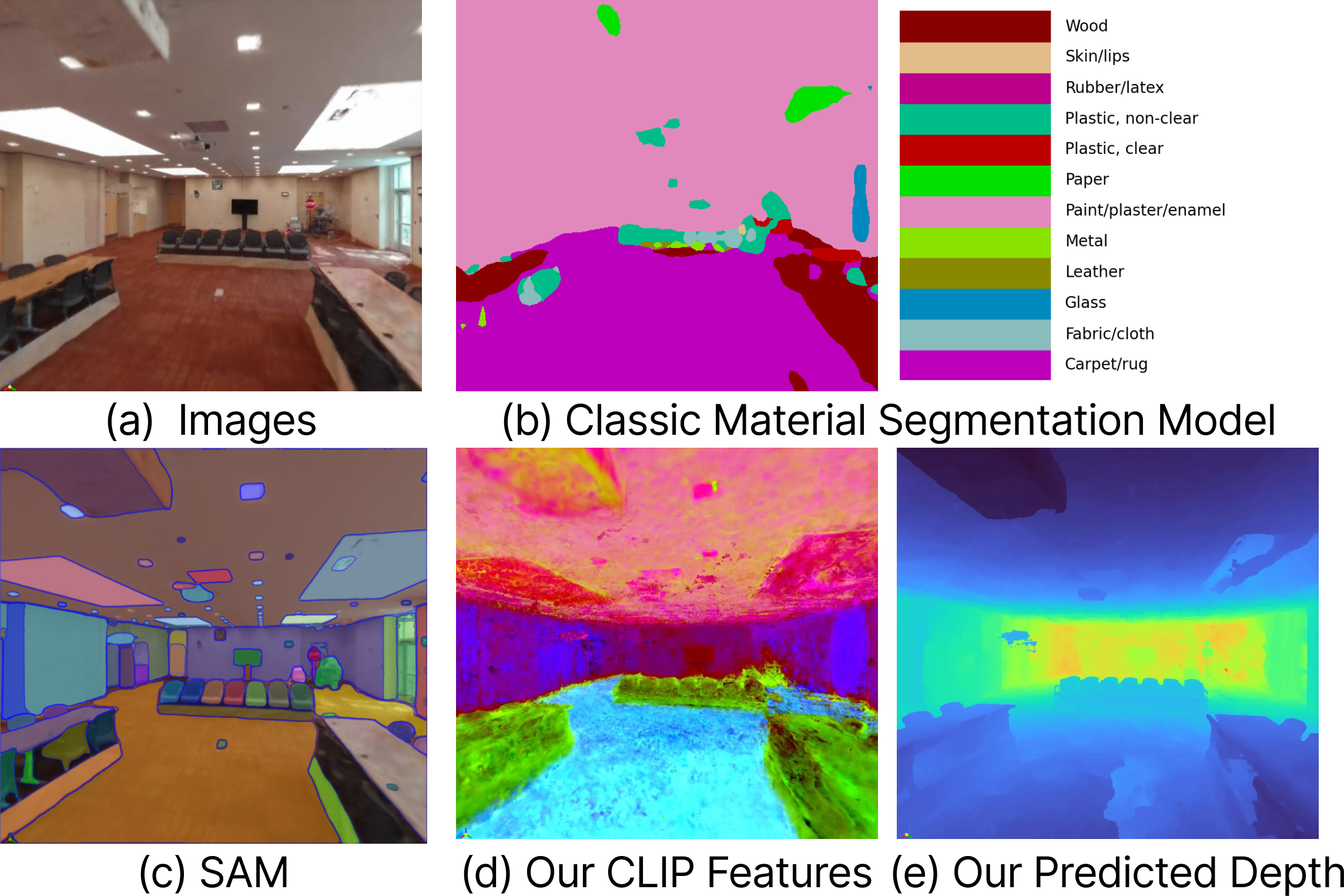}\vspace{-4mm}
    \caption{Comparison of material understanding methods in indoor environments. \textnormal{(a) Input image of a complex indoor scene.
(b) Classic material segmentation models struggle with generalization, producing coarse and inaccurate segmentation that fails to adapt to the diversity of real-world materials.
(c) The Segment Anything Model (SAM)~\cite{cen_segment_2024} offers high-quality instance boundaries but lacks the ability to differentiate materials, limiting its applicability for material-aware tasks.
(d, e) Our vision-guided model extracts rich semantic features using CLIP and simultaneously predicts accurate depth maps, enabling projection onto 3D surfaces.}}\vspace{-7mm}
    \label{fig:sys-segmentation}
\end{figure}

\noindent(3)~\textbf{Differentiable Ray Tracing based Fine Tuning}. Finally, the reconstructed geometry and vision-initialized material maps are fed into a differentiable ray tracer to compute multipath channels. Using a few channel soundings as supervision, the system backpropagates the errors to adjust EM properties, quickly improving accuracy with minimal data.

\vspace{-3mm}
\section{System Design}
\label{s:system_design}

We elaborate on our system design in this section.
\vspace{-3mm}
\subsection{Semantic Feature Distillation}

At the heart of accurate wireless digital twins lies the need to know not only \emph{where} surfaces are, but also \emph{what they are made of}. Unfortunately, classic 2D-to-3D material segmentation pipelines do not generalize well to complex real-world scenes. As illustrated in Fig.~\ref{fig:sys-segmentation}(b), existing material segmentation networks produce coarse, vocabulary-limited masks that often misalign with geometry and fail to capture small objects—errors that quickly propagate in 3D and render such systems (e.g., mmSV~\cite{kamari_mmsv_2023}) ineffective for material-aware channel modeling. Recent advances like the Segment Anything Model (SAM) improve boundary quality (Fig.~\ref{fig:sys-segmentation}(c)) but still lack material awareness: they delineate instances but cannot distinguish wood from glass, or fabric from metal, which is crucial for electromagnetic modeling.

To overcome these limitations, \oursystem leverages large vision–language models (VLMs) such as CLIP to extract open-vocabulary, fine-grained semantic features from raw RGB inputs (Fig.~\ref{fig:sys-segmentation}(d)). Unlike fixed-label segmenters, CLIP embeddings are sensitive to subtle texture and appearance cues, enabling the system to capture material semantics even for uncommon or small-scale objects. Then, we jointly predict geometry cues (depth) alongside semantic features (Fig.~\ref{fig:sys-segmentation}(d, e)), allowing features to be accurately projected into 3D.

\begin{figure}[t!]
    \centering
    \includegraphics[width=1\linewidth]{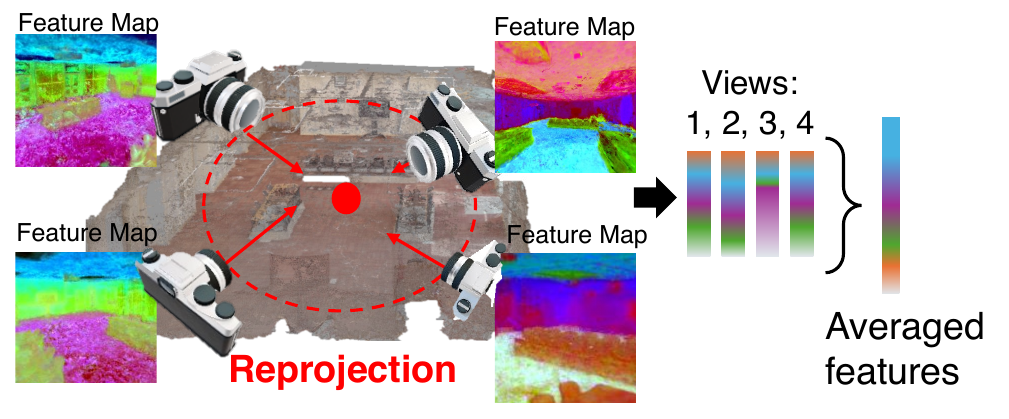}\vspace{-5mm}
    \caption{Feature Re-projection. }\vspace{-7mm}
    \label{fig:sys-projection}
\end{figure}

\noindent\textbf{(1) Multi-View Semantic Embedding.}
A natural idea is to use existing pixel-level VLM models like OpenSeg~\cite{ghiasi_scaling_2022} or LSeg~\cite{li_language-driven_2022} to extract per-pixel material labels and then project them to the 3D point cloud~\cite{engelmann_opennerf_2024,peng_openscene_2023}. These models offer fine-grined scenario understanding. According to our study, they suffer from two key limitations: (1) limited label diversity due to small, fixed training vocabularies, and (2) reduced sensitivity to fine-grained texture queries like materials. Our experiments reveal that such models are insufficient for predicting nuanced radio-material behavior. One might instead try to extract CLIP features directly for each 3D point. However, this method is still problematic -- CLIP's design for global image-level embeddings means that it does not provide the fine-grained, pixel-level detail needed. As a result, sampling from a projected 2D location usually yields noisy and spatially ambiguous data, a problem that is particularly acute for small objects~\cite{kerr_lerf_2023}.

\begin{figure}[t!]
    \centering
    \includegraphics[width=0.6\linewidth]{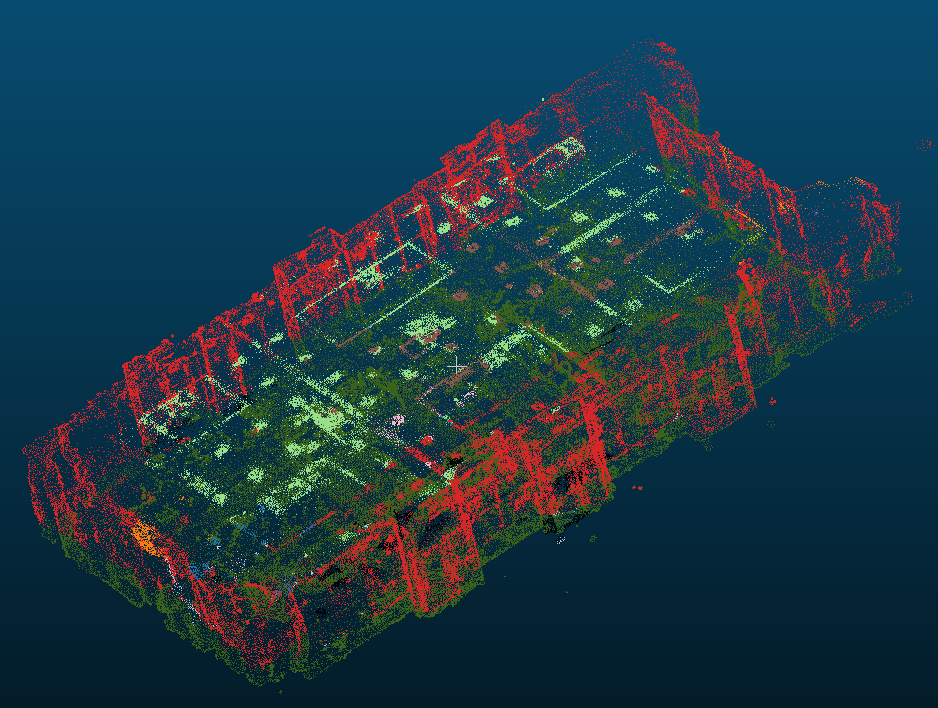}\vspace{-3mm}
    \caption{3D Semantic Field. \textnormal{The colors encode semantic features, with different hues indicating distinct material categories across the scene.}}\vspace{-5mm}
    \label{fig:3d-semantic}
\end{figure}

We instead train a neural radiance field (NeRF) model $\mathcal{E}^{3D}$ as our semantic lifting engine (Fig.~\ref{fig:sys-overview}). Unlike a standard NeRF that use a nerual network $\mathcal{E}^{3D}$ to map 3D positions $(x,y,z)$ and viewing directions $(\theta, \phi)$ to color and density for novel view synthesis, $\mathcal{E}^{3D}$ is extended to also output a semantic feature vector at each voxel. During training, we render rays through $\mathcal{E}^{3D}$ and jointly supervise the predicted color, density, and semantics. Semantic supervision is enforced by aligning the per-ray features with local CLIP embeddings extracted from the corresponding 2D image views. Specifically, given a set of $N$ pose-calibrated RGB images $\mathcal{I}=\{I_i\}_{i=1}^N$, we first extract high-dimensional vision-language embeddings $\mathbf{E}_i \in \mathbb{R}^{C}$ from each image using pretrained vision language models $\mathcal{E}^{2D}$ (CLIP~\cite{wu_how_2022} and DINOv2~\cite{oquab_dinov2_2024}),  where $C$ is the embedding channel dimension ($C=768$ in CLIP Base and DINOv2 Base). In this way, every pixel in the training images is associated with a corresponding semantic embedding, which we treat as ground-truth supervision for the underlying 3D samples along its camera ray. To incorporate this into the NeRF architecture, we augment the MLP with an additional output head that predicts a semantic feature vector at each 3D location. During training, these per-sample features are accumulated via standard volume rendering along the ray to produce a rendered semantic embedding for each pixel. We then supervise this rendered embedding against the pixel-aligned CLIP ground truth using a cosine similarity loss.

To further strengthen semantic supervision, we adopt the multi-scale strategy of LERF~\cite{kerr_lerf_2023}. Instead of relying solely on single-pixel embeddings, we construct a feature pyramid by cropping each training image into overlapping windows at multiple scales and extracting CLIP features from each crop. For a given training ray passing through pixel $(u,v)$, we then interpolate within this pyramid to obtain a local CLIP embedding $\mathbf{f}_{j}^{\text{CLIP}}$ for the 3D point $\mathbf{p}_j$ along the ray. This provides scale-aware supervision, allowing the NeRF to capture both coarse context and fine-grained cues , and ensures that the learned semantic field remains consistent across views and resolutions.

\noindent\textbf{(2) Lifting 2D Semantics to 3D.}
Having trained a NeRF-based semantic extractor with multi-scale CLIP supervision, the next step is to consolidate these per-ray predictions into a fine-grained 3D semantic field.

Specifically, as shown in Fig.~\ref{fig:sys-projection}, for any 3D surface point $\mathbf{p}_j \in \mathbb{R}^3$, we identify all camera views in which the point is visible. $
\mathcal{I}(\mathbf{p}_j)=\{\,I_i\in\mathcal{I}\mid\mathbf{p}_j\text{ is visible in }I_i\}
$. For each such view, we project the 3D point into the image using the intrinsic and extrinsic parameters:
\begin{equation}\small
(u, v) = \pi(I_i, E_i, \mathbf{p}_j),
\end{equation}
where $\pi$ is the camera projection function and $(u, v)$ is the pixel coordinate of the projected point. $I_i$ and $E_i$ are the camera intrinstic and extrinsic matrix.
We then query the trained NeRF $\mathcal{E}^{3D}$ and do volumetric rending along the ray corresponding to pixel $(u,v)$ in image $I_i$, obtaining the predicted semantic feature vector:
\begin{equation}\small
\mathbf{e}_{i,j} = \mathcal{E}^{3D}(\mathbf{p}_j \mid I_i, E_i).
\end{equation}
This allows us to retrieve view-specific semantic predictions that capture both visual context and geometric consistency.
To obtain a robust and view-invariant semantic representation for $\mathbf{p}_j$, we average the predicted features across all $N$ views as below:
\begin{equation}\small
\hat{\mathbf{f}}_j^{3D} = \frac{1}{N} \sum_{i=1}^{N} \mathbf{e}_{i, j}.
\end{equation}
This fused descriptor $\hat{\mathbf{f}}_j^{3D}$ represents an open-vocabulary, CLIP-aligned semantic embedding anchored at a 3D point. It reflects the radiance field’s interpretation of how that point appears from multiple perspectives, enriched with high-level material cues learned through CLIP supervision. The resulting 3D semantic field supports  electromagnetic property prediction, providing a dense, explainable, and generalizable material understanding of the scene as shown in Fig.~\ref{fig:3d-semantic}.

\begin{figure}[t!]
    \centering
    \includegraphics[width=0.95\linewidth]{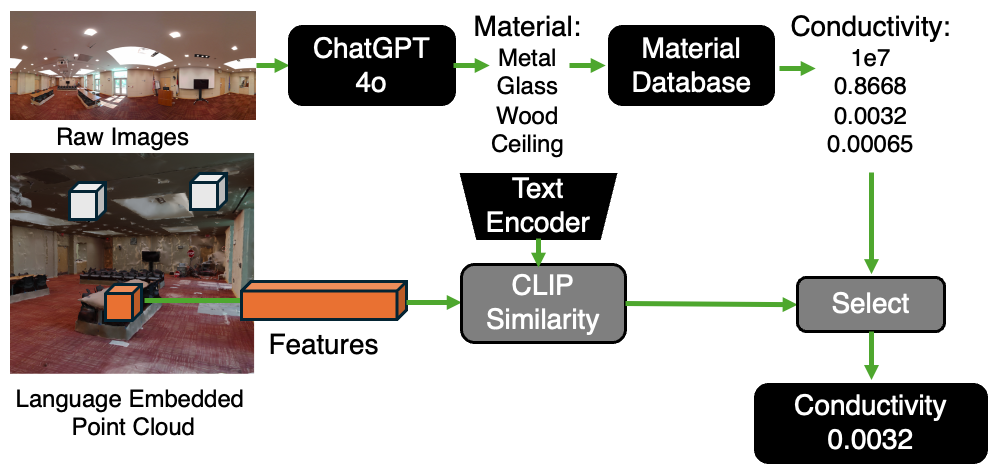}\vspace{-3mm}
    \caption{Pipeline for radio material identification.}
    \label{fig:sys-material-pipeline}
    \vspace{-5mm}
\end{figure}

\subsection{Vision to EM Parameter Conversion}\label{sec:vision-em}
While the NeRF-predicted semantic features $\hat{\mathbf{f}}_j^{3D}$ encode rich visual and contextual cues and can be used for general material identification, they lack direct physical meaning in the electromagnetic (EM) domain. To bridge this semantic-to-physical gap, we translate these vision-language embeddings into radio-relevant material properties—specifically, complex permittivity and conductivity, as shown in Fig.~\ref{fig:sys-material-pipeline}.

\textbf{Step 1: Extracting Material Candidates with VLM.}
Given raw images and 3D scene context, we first query a powerful pretrained vision–language model (e.g., ChatGPT-4o) with a structured prompt (Fig.~\ref{fig:sys-material-prompt}). The model will output a concise JSON list of plausible material categories (e.g., \texttt{["wood","glass","metal","carpet"]}), which serves as the open-set search space for subsequent matching.

\textbf{Step 2: Encoding Text Descriptions.}
Each candidate material name is encoded using the same vision–language embedding space (e.g., CLIP's text encoder), producing a set of normalized textual embeddings:
\begin{equation}
\mathcal{T} = \{\mathbf{t}_1, \mathbf{t}_2, \dots, \mathbf{t}_M\},
\end{equation}
where each $\mathbf{t}_m$ corresponds to a material candidate.
\begin{figure}[t!]
    \centering
    \includegraphics[width=0.5\linewidth]{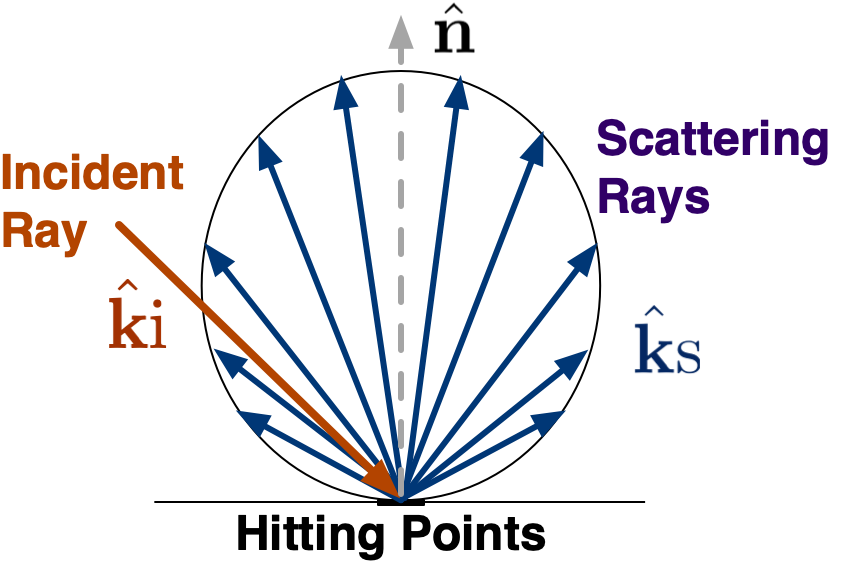}\vspace{-5mm}
    \caption{Lambertian scattering model. \textnormal{An incident ray striking a perfectly diffuse surface results in scattered rays distributed uniformly over the hemisphere} }
    \label{fig:scattering}\vspace{-5mm}
\end{figure}

\textbf{Step 3: Feature Similarity Matching.}
For each 3D point $\mathbf{p}_j$, we compute the cosine similarity between the predicted semantic feature $\hat{\mathbf{f}}_j^{3D}$ and all text embeddings $\mathbf{t}_m$:
\begin{equation}
s_{j,m} = \cos(\hat{\mathbf{f}}_j^{3D}, \mathbf{t}_m).
\end{equation}
The material candidate with the highest similarity score is selected. To improve robustness, we represent each material using a set of synonymous or related word embeddings (e.g., ``concrete,'' ``cement,'' ``pavement''), and select the candidate with the maximum aggregated similarity.

\textbf{Step 4: Material Database Lookup.}
Finally, we query a curated electromagnetic material database using the selected label $\hat{m}_j$. This database maps textual material categories to their corresponding physical parameters, including relative permittivity $\epsilon_r$, conductivity $\sigma$. We initialize this database following the ITU standard material definitions~\cite{series2015effects}, and extend it with additional entries as needed. For materials not present in the database, we leverage a large language model (e.g., GPT-5) to infer radio parameters based on the images, which are then refined through subsequent calibration.

\noindent\textbf{Discussion.} The VLM is essential to our design because it provides open-set, context-aware material reasoning that cannot be achieved by CLIP-similarity alone. It extracts point-level material categories and coarse surface attributes (e.g., wood, metal, glass, roughness), allowing us to assign plausible permittivity, conductivity, and scattering priors through a curated EM database. Although these priors are not perfect, they offer a far stronger initialization than deep-learning baselines, enabling over a 10$\times$ reduction in calibration measurements while preserving multipath accuracy. Moreover, as a large pretrained model with open-vocabulary recognition, the VLM reliably re-identifies materials across different environments, making cross-scene reuse and rapid adaptation possible and positioning it as a critical component of \oursystem.

\begin{table}[t!]
\centering \small
\caption{Scattering Coefficient of Typical Materials.
\normalfont{We list a few examples for illustration.}}\label{tab:roughness_scattering}
\vspace{-3mm}
\begin{tabular}{|l|c|c|}
\hline
\textbf{Material} & \textbf{Ra (µm)} & \textbf{Scattering Coeff. $S$} \\
\hline
Wall Plasterboard & 2.309 & 0.07--0.1 \\
Chipboard         & 2.695 & 0.1--0.2 \\
Cardboard         & 3.252 & 0.1--0.2 \\
Ceiling Plasterboard & 11.85 & 0.2--0.4 \\
Brick             & 14.68 & 0.3--0.5 \\
\hline
\end{tabular}\vspace{-5mm}
\end{table}

\subsection{Vision to Scattering Modeling}

According to previous research~\cite{mi_measurement-based_2024}, scattering rays contribute about 30~\% for the whole mmwave channel. To accurately characterize diffuse scattering in mmWave environments, we begin with a general scattering model, which distinguishes between specular and diffuse reflections. The scattering coefficient $S \in [0, 1]$ quantifies the proportion of energy that is diffusely scattered, with the remaining energy being specularly reflected. This relationship is expressed through the reflection reduction factor $R = \sqrt{1 - S^2}$. To capture the spatial directivity of diffuse reflections, we adopt the Lambertian scattering model~\cite{hoydis_sionna_2023} as shown in Fig.~\ref{fig:scattering}, which assumes energy is scattered with a cosine-weighted distribution centered around the surface normal. Mathematically, this behavior is described by the bidirectional scattering distribution function (BSDF) for Lambertian surfaces as:
\begin{equation}f^\text{Lambert}\text{s}\left(\hat{\mathbf{k}}\text{i}, \hat{\mathbf{k}}\text{s}, \hat{\mathbf{n}}\right) = \frac{\hat{\mathbf{n}}^\mathsf{T} \hat{\mathbf{k}}\text{s} }{\pi} = \frac{\cos(\theta_s)}{\pi}
\end{equation}
Here, $\hat{\mathbf{k}}\text{i}$ and $\hat{\mathbf{k}}\text{s}$ denote the unit incident and scattering directions, respectively, and $\hat{\mathbf{n}}$ is the unit surface normal. The function depends only on the angle $\theta_s$ between the surface normal and the scattered direction, emphasizing the physically realistic notion that light is more likely to scatter close to the normal direction than tangentially.

\begin{figure}[t!]
    \centering
    \includegraphics[width=\linewidth]{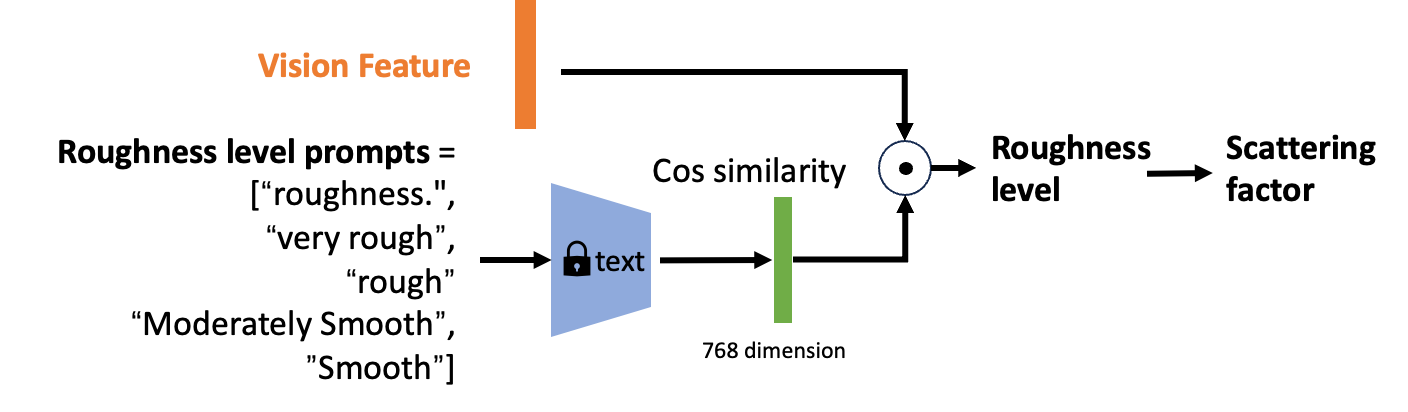}\vspace{-5mm}
    \caption{Scattering coefficient prediction using vision-language similarity. \textnormal{Vision features extracted from material images are compared with text embeddings of roughness-related prompts to infer scattering levels.}}
    \label{fig:scattering-prediction}\vspace{-5mm}
\end{figure}

To estimate the scattering coefficient $S$ of indoor materials in our vision-based radio twin system, we leverage prior empirical measurements from mmWave channel studies (Table~\ref{tab:roughness_scattering}). These studies demonstrate a clear correlation between material roughness—quantified using the measured average arithmetic roughness parameter $Ra$—and the associated diffuse scattering behavior under the Lambertian model. Notably, smoother materials like wall plasterboard exhibit lower scattering coefficients (e.g., $S = 0.07 - 0.1$), whereas rougher surfaces like brick show higher values (up to $S = 0.5$). In practice, directly global CLIP matching (Fig.~\ref{fig:scattering-prediction}) which may struggle with fine textures. We first decompose the material surface into local patches. We then leverage a VLM (e.g., GPT-4o) to perform zero-shot classification on these patches using the discrete scattering level prompt and then use the corresponding entry in the table to obtain the scattering-coefficient prior.

Fig.~\ref{fig:scattering-material} shows examples of indoor surfaces with increasing roughness—plasterboard, cardboard, and brick—where the visual texture correlates with $Ra$ roughness values. By leveraging vision–language models, we can readily distinguish such texture levels and map them to scattering coefficients, providing a practical basis for perceptually aware wireless digital twins in mmWave signal modeling.
This motivates our approach: by using pre-trained vision-language models, we extract visual features from images of materials to infer surface roughness levels in Table.~\ref{tab:roughness_scattering}. These inferred roughness categories can then be directly mapped to scattering coefficient intervals, enabling vision-guided estimation of diffuse scattering properties for enhanced spatial modeling in our mmWave radio twin system.

\vspace{-5mm}
\subsection{Few-Shot Calibration}

\begin{figure}[t!]
    \centering
    \includegraphics[width=\linewidth]{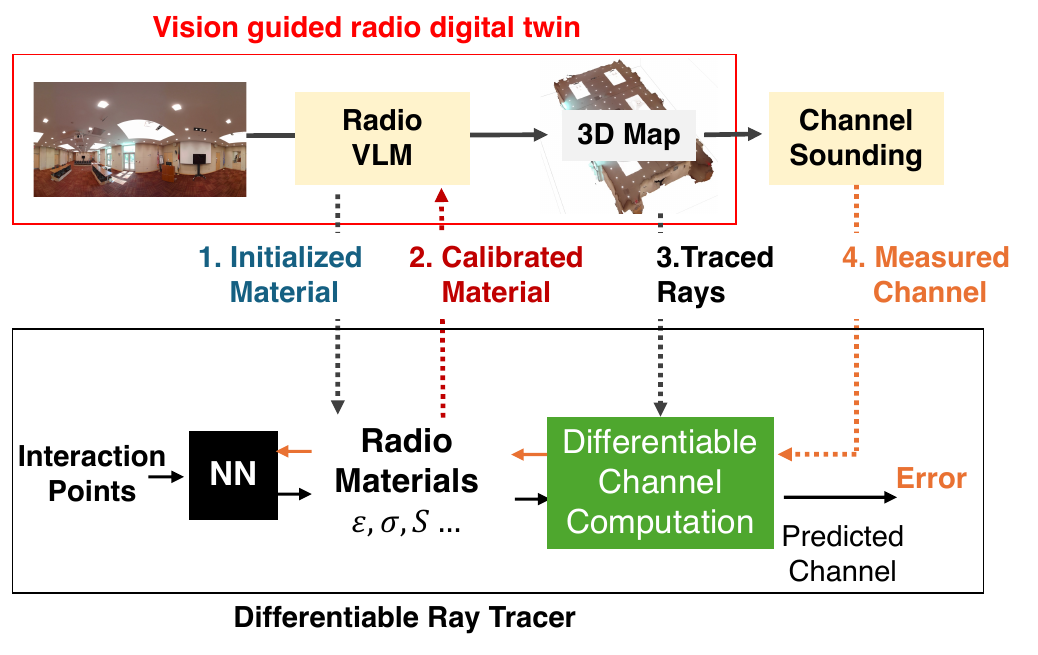}\vspace{-5mm}
    \caption{Few‑Shot Vision‑Guided Radio Digital‑Twin Training. \textnormal{After running the vision based radio twin, a few channel sounding probes anchor this twin, and differentiable path tracing compares predicted channel with measurements. The loss is back‑propagated to calibrate the object material.}}\vspace{-7mm}
    \label{fig:sys-fewshot}
\end{figure}

As shown in Fig.~\ref{fig:sys-fewshot}, the vision pipeline provides reliable scene geometry and material \emph{priors}, but some mismatch is inevitable. For example, surface color can mislead semantic classification, roughness can only be inferred indirectly, and centimetre-scale drift may persist in the reconstructed NeRF mesh. To close these gaps, we introduce a \emph{few-shot} calibration step that uses differentiable ray tracing and requires few channel soundings per scene.

\noindent$\bullet$ \textbf{Differentiable ray-tracing core.}
We build on Sionna RT~\cite{hoydis_sionna_2023}, which can render the complex channel impulse response
\(\hat{\mathbf{h}}(\mathbf{p};\boldsymbol{\Omega})\) for a transmitter–receiver pair at position \(\mathbf{p}\), where $\Omega$ is the learned radio environment. Importantly, the renderer also exposes analytic gradients \(\nabla_{\boldsymbol{\Omega}}\hat{\mathbf{h}}\) with respect to a parameter vector \(\boldsymbol{\Omega}\).
In our case, \(\boldsymbol{\Omega}\) contains \textbf{trainable material parameters} for each surface patch: conductivity \(\sigma\), relative permittivity \(\varepsilon_r\), scattering coefficient \(S\).

To ensure stable optimization and physically valid values, we represent each surface \(o\) with a learnable embedding \(\mathbf{v}^{(o)}\in\mathbb{R}^{L}\).
Four linear projections convert this embedding into physical parameters, enforced through exponential or sigmoid transforms:
\begin{align}\footnotesize
  \sigma      &= \exp\!\bigl(\mathbf{v}^{(o)\!\top}\mathbf{w}_1\bigr), \\
  \varepsilon_r &= 1 + \exp\!\bigl(\mathbf{v}^{(o)\!\top}\mathbf{w}_2\bigr), \\
  S           &= \mathrm{sigmoid}\!\bigl(\mathbf{v}^{(o)\!\top}\mathbf{w}_3\bigr)
\end{align}
These parameterizations guarantee \(\sigma > 0\), \(\varepsilon_r \geq 1\), and \(S \in [0,1]\). Embeddings are initialized from the vision-estimated materials and then refined by gradient descent.

\begin{figure*}[t!]
    \centering
        \begin{subfigure}[b]{0.24\linewidth}
        \centering
        \includegraphics[width=\linewidth]{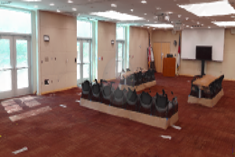}
        \caption{Indoor: Conference Room}
        \label{fig:exp-conference}
    \end{subfigure}
    \hfill
    \begin{subfigure}[b]{0.24\linewidth}
        \centering
        \includegraphics[width=\linewidth]{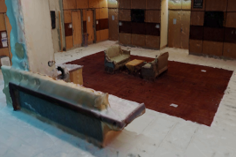}
        \caption{Indoor: Lobby}
        \label{fig:indoor-ray}
    \end{subfigure}
    \hfill
    \begin{subfigure}[b]{0.23\linewidth}
        \centering
        \includegraphics[width=\linewidth]{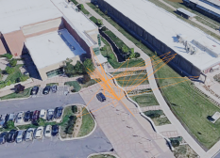}
        \caption{Outdoor Scenarios.}
        \label{fig:outdoor-ray}
    \end{subfigure}
    \hfill
    \begin{subfigure}[b]{0.26\linewidth}
        \centering
        \includegraphics[width=\linewidth]{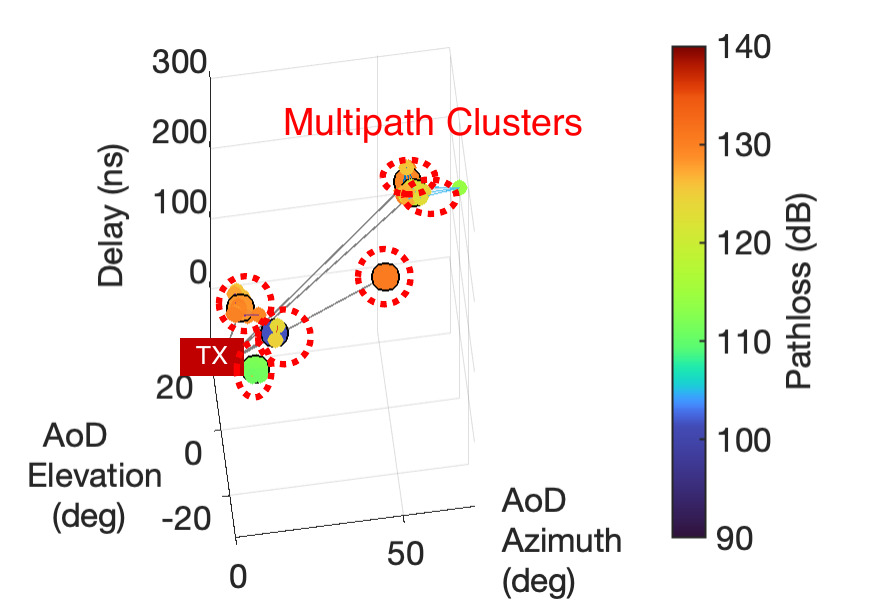}
        \caption{Multipath Profile}
        \label{fig:exp-exp-multipath}
    \end{subfigure}\vspace{-4mm}
    \caption{Evaluation Scenarios. \textnormal{(a) Indoor conference room with tables and chairs; (b) Indoor lobby space with sofas, tables, and glass partitions used for millimeter-wave measurements; (c) Outdoor scenario besides a parking lot; (d) Measured multipath profile depicting received power across individual paths.}}\vspace{-5mm}
    \label{fig:scenarios}
\end{figure*}

\noindent$\bullet$ \textbf{Initialization from vision priors.}
Before calibration, each surface patch must be assigned a reasonable starting point for its material parameters. We leverage the estimates provided by the vision-guided pipeline: the VLM supplies a candidate material label (e.g., wood, glass, concrete), which is then mapped to a table of nominal radio parameters \((\sigma_0,\varepsilon_{r0},S_0)\) drawn from standard EM property databases. These values serve as prior “guesses” for the calibration process. To encode them into our trainable embeddings, we solve the inverse of the parameterization equations. In this way, each embedding \(\mathbf{v}^{(o)}\) is initialized such that its read-out exactly reproduces the vision-estimated parameters at iteration zero. Subsequent gradient descent then fine-tunes these values toward the measured channel responses. This warm start avoids poor local minima and ensures that the few-shot calibration acts as a refinement of physically meaningful priors rather than a blind search.

\noindent$\bullet$ \textbf{Measurement features and loss.}
Each mmWave sounding snapshot provides a measured impulse response \(\mathbf{h}_d\). Instead of aligning raw CIR samples, we extract two robust, time-offset agnostic features:
the received power \\$P_d=\sum_{\ell}\lvert h_d[\ell]\rvert^{2}$, where $h_d$ is the path strength and the delay spread
\begin{equation}\footnotesize
\tau_d=\sqrt{\sum_{\ell}\Bigl(\tfrac{\ell-\bar{\tau}_d}{W}\Bigr)^{2}
               \tfrac{\lvert h_d[\ell]\rvert^{2}}{P_d}}
\end{equation}
where \(\bar{\tau}_d\) is the mean delay and \(W\) the sample rate.

The tracer produces corresponding predictions \(\hat{P}_d,\hat{\tau}_d\). Calibration minimizes the batch-averaged symmetric mean absolute percentage error (SMAPE):
\begin{equation}\footnotesize
\mathcal{L}=\frac{1}{B}\sum_{d=1}^{B}\Bigl[
\lambda_{P}\frac{\lvert P_d-\hat{P}_d\rvert}{P_d+\hat{P}_d}
+
\lambda_{\tau}\frac{\lvert \tau_d-\hat{\tau}_d\rvert}{\tau_d+\hat{\tau}_d}
\Bigr]
\end{equation}
where \(\lambda_{P},\lambda_{\tau}\) balance the power and delay-spread terms. An exponential moving average scale factor compensates for the unknown absolute power reference of the channel sounder. In short, this calibration loop leverages differentiable physics to fine-tune vision-derived material priors with only a few probe soundings, ensuring the digital twin remains accurate with minimal data collection.
\vspace{-3mm}
\subsection{Building 3D Map and Ray Tracing}

To support ray tracing and channel modelling, we derive a mesh-based 3D map by querying the NeRF model with virtual camera rays to obtain per-pixel depth estimates, which are back-projected into a dense point cloud. We then construct a clean mesh by applying standard point-cloud filtering and Poisson surface reconstruction. The full processing pipeline is described in Appendix~\ref{appendix:pcd}. The impact of geometry error is summarized in Appendix~\ref{appx:geo-error}. We then import the 3D mesh into the ray tracer and use shoot and bounding method to identify the propogation paths. For each transmitter–receiver path set $\mathcal{P}$, we compute per-path gain
\[
  G_p(f) = \prod_{j\in p} M(\sigma_j,\varepsilon_j,s_j;f),
\]
where $M(\cdot)$ models reflection and transmission at point $j$.  Summing over $p$ produces the channel response
\[
  \hat{h}(f)=\sum_{p\in\mathcal{P}}G_p(f),
\]
Once we get a hitting point, we can query the NeRF neural network to get the vision features and get the radio parameters using the pipeline discussed in the Section~\ref{sec:vision-em}.

\vspace{-3mm}
\subsection{Dealing with Dynamics}
Environmental dynamics—such as moving pedestrians, vehicles, or shifting furniture—pose a fundamental challenge to accurate wireless channel modelling, as mmWave links are highly sensitive to transient blockages and mobility that can cause severe throughput fluctuations~\cite{kamari_mmsv_2023}. To address this, our system continuously fuses visual sensing with lightweight radio calibration: pervasive cameras (e.g., CCTV or AR glasses) detect and localize dynamic objects in real time, and vision-based tracking updates only the affected regions of the 3D mesh, after which the differentiable ray tracer recomputes altered multipath paths. For example, when a person walks through a link, a structure-from-motion tracker inserts a proxy mesh of the moving body with an embedding initialized from the latest camera frame. In this way, the system incrementally adapts to dynamics, achieving resilience to transient obstructions without requiring full retraining or dense site-specific surveys.

\begin{figure*}[t!]
  \centering
  \begin{subfigure}{.2\textwidth}
    \includegraphics[width=\linewidth]{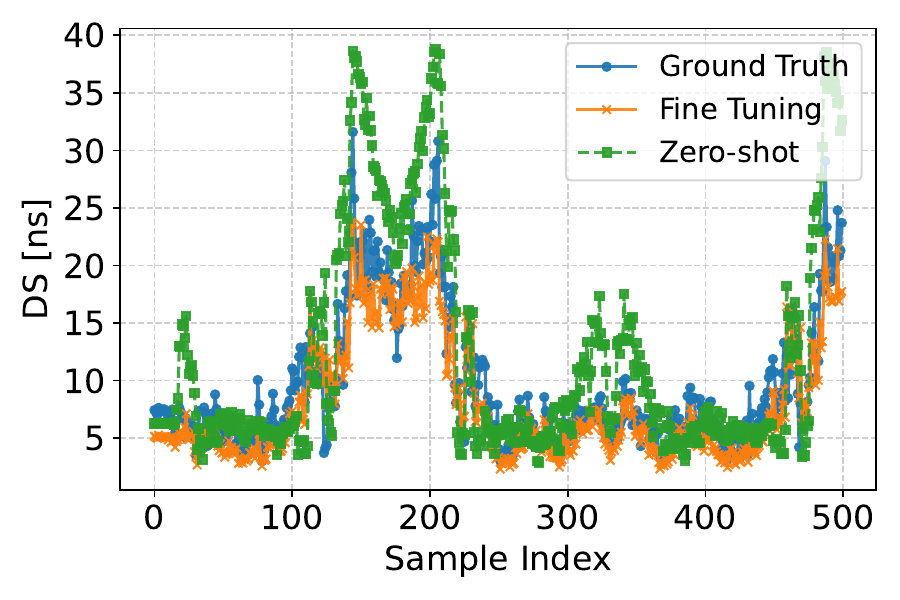}\vspace{-2mm}
    \caption{$\tau$ RMS}
    \label{fig:rms}
  \end{subfigure}\hfill
    \begin{subfigure}{.2\textwidth}
    \includegraphics[width=\linewidth]{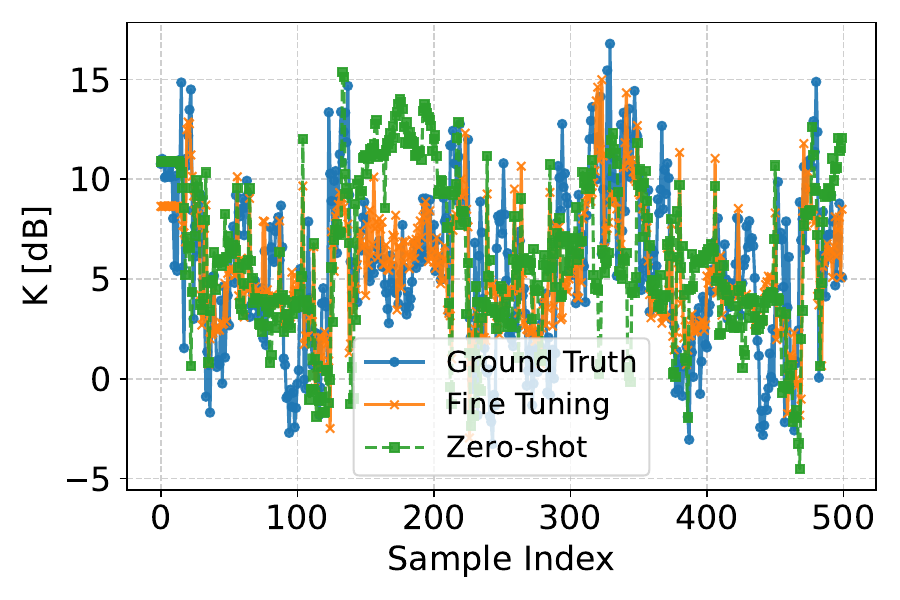}\vspace{-2mm}
    \caption{K factor}
    \label{fig:k-factor}
  \end{subfigure}\hfill
  \begin{subfigure}{.2\textwidth}
    \includegraphics[width=\linewidth]{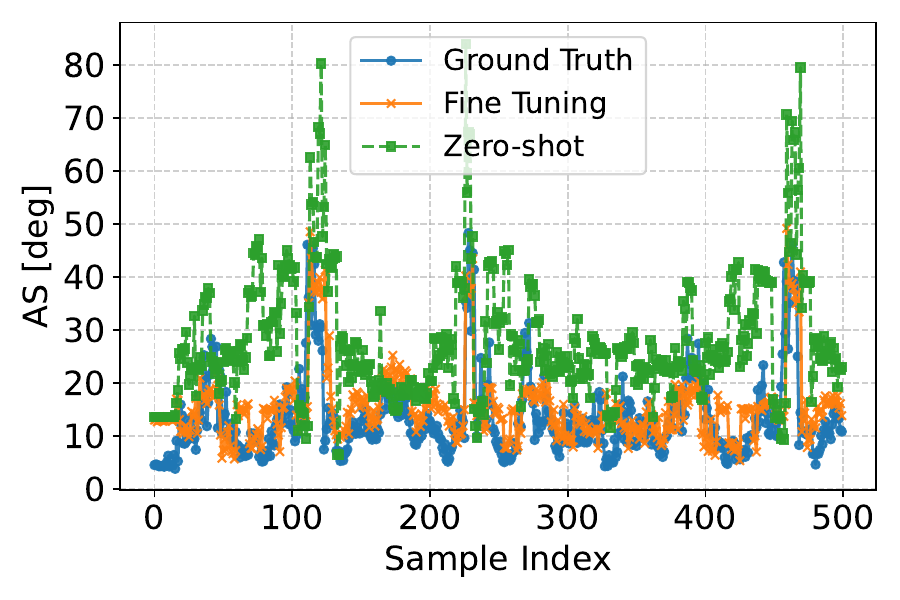}\vspace{-2mm}
    \caption{Angular Spread}
    \label{fig:as}
  \end{subfigure}\hfill
    \begin{subfigure}{.2\textwidth}
    \includegraphics[width=\linewidth]{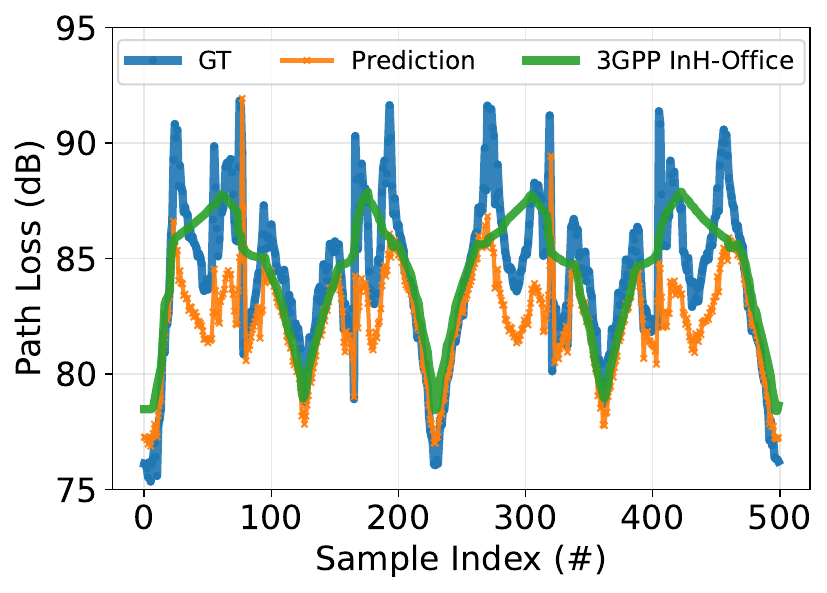}\vspace{-2mm}
    \caption{Path Loss}
    \label{fig:pl-1}
  \end{subfigure}\hfill
      \begin{subfigure}{.2\textwidth}
    \includegraphics[width=\linewidth]{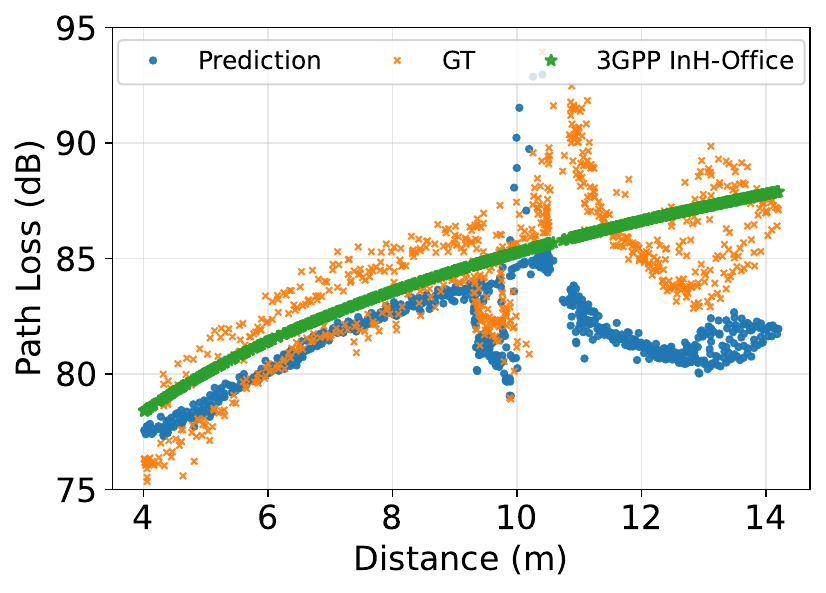}\vspace{-2mm}
    \caption{Path Loss v.s. Distance}
    \label{fig:pl-2}
  \end{subfigure}\hfill
  \vspace{-5mm}
  \caption{Snapshot-level performance of the proposed model on the four canonical channel-metrics.}\vspace{-5mm}
  \label{fig:metric-results}
\end{figure*}
\vspace{-3mm}
\section{Implementation}
\label{sec:implementation}

In this section, we describe the evaluation dataset (\S\ref{ss:dataset}) and model training (\S\ref{ss:model_training}).
\vspace{-3mm}
\subsection{Dataset Preparation}
\label{ss:dataset}

We evaluate \oursystem\ using a series of context-aware datasets collected with a channel sounder~\cite{gentile_context-aware_2024}, which integrates a 60.5~GHz RF transceiver (2~GHz bandwidth) with an 8-horn Tx and 16-horn Rx antenna array, an iSTAR Pulsar panoramic camera (11000 × 5500 spherical images, NCTech), and an OSO-128 LiDAR (0.17$^\circ$ azimuth resolution, Ouster). The Tx, mounted at a height of 2.5~m, transmits a 2047-bit PN sequence, while the Rx—mounted on a mobile robot moving at (0.2-0.3)~m/s—collects multipath profiles every (10–15)~cm. The demo multipath profile measurement is listed in Appendix~\ref{appx:multipath-profile}.As shown in Fig.~\ref{fig:scenarios}, measurements were conducted in (1) a 20~m × 10~m lecture room~\cite{mi_measurement-based_2024} (4956 samples, Line of Sight (LoS)-dominated) furnished with tables and chairs; (2) a 20~m × 15~m lobby (5100 samples, \textbf{including non-LoS (NLoS) paths} as shown in Fig.~\ref{fig:exp3-spatial}) containing sofas, tables, walls, and glass panels~\cite{gentile_context-aware_2024}; and (3) an outdoor scenario illustrated in Fig.~\ref{fig:outdoor-ray}. Data synchronization was maintained using Rubidium clocks and control markers. For each multipath component, we record the AoA and AoD (azimuth and elevation), delay, and path loss, as shown in Fig.~\ref{fig:exp-exp-multipath}. To evaluate the impact of human motion, we use the dataset shown in~\cite{mukherjee_scalable_2022} where a human subject moves in between a Tx and an Rx. To obtain accurate camera poses and train NeRF-based vision models, we process raw images and LiDAR point clouds in Agisoft Metashape to reconstruct the scene, calibrate camera poses, and then export Nerfstudio-compatible datasets. For outdoor scenarios, we use Google Earth Studio to generate virtual views at specified camera poses and then construct Nerfstudio-compatible datasets.

\vspace{-3mm}
\subsection{Model Training}
\label{ss:model_training}

We implement our model based on nerfstudio~\cite{tancik_nerfstudio_2023} and use around 100 panoramic images as input for reconstructing each scene. The model runs on a server equipped with an NVIDIA A6000 GPU. We launch ray tracing based on Sionna ray tracer v0.19.2~\cite{hoydis_sionna_2023}. In each ray tracing, we launch 4e6 rays and the maximum depth for one ray is 5. The probability with which a scattered path is kept is set to 1e-3.
To ensure alignment between the path strength in ray tracer and the measured path strength, we introduce a global scaling factor. This factor is calibrated per scenario by matching the mean predicted LoS path strength to the corresponding measured LoS magnitude.

\section{Evaluation}
\label{sec:evaluation}

We evaluate {\oursystem} across indoor and outdoor scenes to assess channel estimation accuracy, data efficiency, and adaptability. Specifically, we examine its performance against baselines on standard channel metrics, study the benefits of vision-guided initialization and few-shot calibration, and analyze its robustness under dynamic scene changes.
\vspace{-3mm}
\subsection{Metrics and Baselines}\label{sec:metrics}
We define a \emph{snapshot} as a single channel measurement at a fixed transmitter–receiver configuration, capturing all multipath components (MPCs) for that link. To obtain a fine-grained view of estimation accuracy, we consider two complementary groups of metrics.
\textbf{Snapshot-level metrics} (M1 – M4) quantify the aggregate behaviour across all MPCs in a snapshot.
\textbf{Cluster-level metrics} (C1–C2) further analyze the internal multipath structure of a snapshot by grouping MPCs into delay–angle clusters and measuring how well each predicted cluster matches its ground-truth counterpart.

\begin{figure}[t!]
  \centering
    \includegraphics[width=0.85\linewidth]{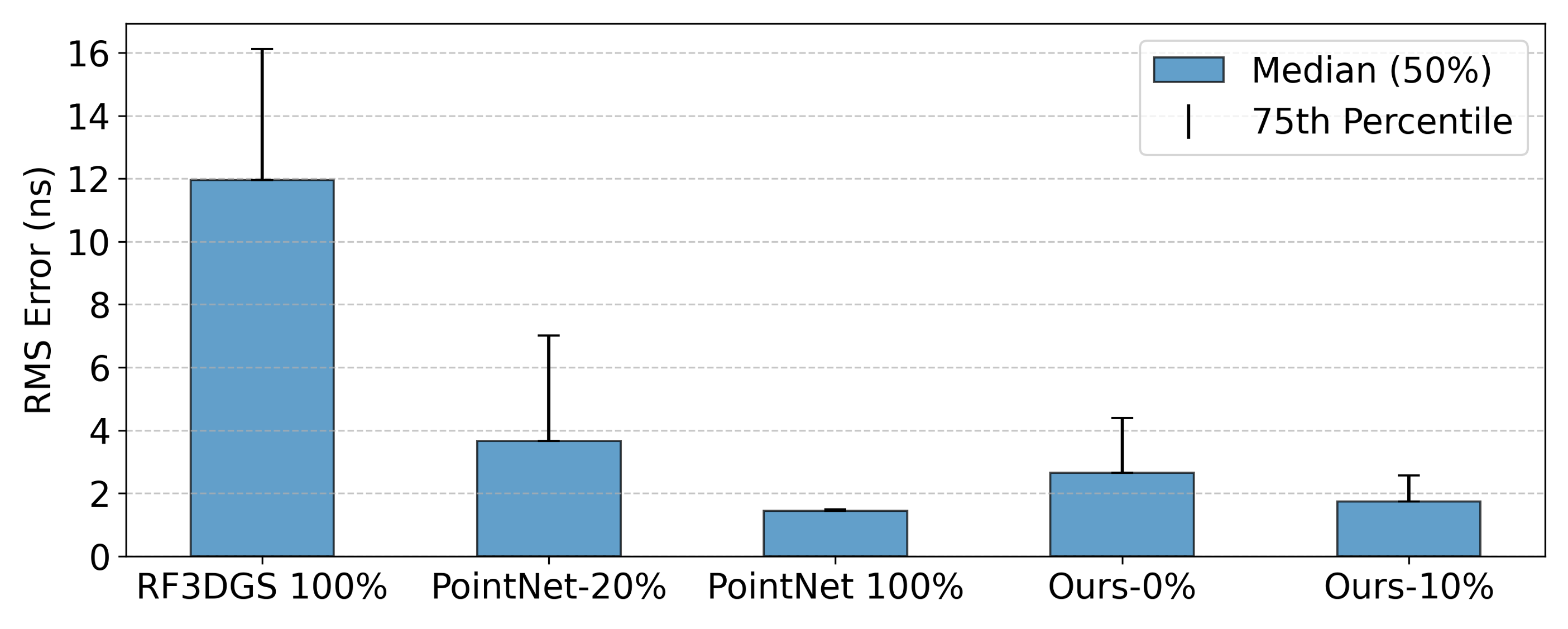}\vspace{-5mm}
  \caption{Compared with SoTA Solution. \textnormal{Delay Spread.}}\vspace{-5mm}
  \label{fig:exp-sota}
\end{figure}

\noindent$\bullet$ \textbf{Snapshot-level metrics (M1–M4)}. We use four standard metrics to capture large-scale and small-scale behavior within each snapshot as shown in Table~\ref{tab:snapshot-metrics}.
\textbf{Path Loss (PL)} quantifies the total received signal attenuation across all multipath components (MPCs).
\textbf{RMS Delay Spread} measures the temporal dispersion of the channel, indicating how much multipath broadens the received signal in time.
\textbf{Angular Spread (AS)} captures the spatial dispersion of multipath in the azimuth domain, reflecting the diversity of arrival directions.
Finally, the \textbf{Rician K-factor} characterizes the power ratio between the line-of-sight path and the aggregate of all scattered MPCs, describing the dominance of deterministic versus diffuse components. These metrics provide a holistic view of channel  per snapshot, while the cluster-level metrics give fine-grained insight into how well the model localizes and weights individual multipath clusters.

\noindent$\bullet$ \textbf{Baselines.}  We compare \oursystem\ against three representative approaches:
(i) the 3GPP path loss model~\cite{3gpp_3gpp_2022} as a standard empirical baseline,
(ii) the PointNet-based deep learning predictor~\cite{narayanan_lumos5g_2020}, which learns channel characteristics directly from extensive site-specific RF measurements.  (iii) RF-3DGS~\cite{zhang_rf-3dgs_2025}, a vision-based neural ray tracing that relies solely on visual geometry priors and lacks any physics-based propagation modeling.

\begin{table}[b!]
\centering\vspace{-5mm}
\caption{Snapshot-level metrics (M1–M4).}\vspace{-3mm}
\label{tab:snapshot-metrics}
\begin{threeparttable}
\footnotesize
\renewcommand{\arraystretch}{1.2}
\begin{tabular}{p{3.2cm}p{4cm}}
\toprule
\textbf{Metric} & \textbf{Definition} \\
\midrule
Path Loss (PL) &
\(\mathrm{PL}= -10\log_{10}\Bigl(\sum_{i=1}^{N} P_i\Bigr)\) \\
RMS Delay Spread (\(\tau_{\mathrm{RMS}}\)) &
\(\tau_{\mathrm{RMS}}
= \sqrt{\tfrac{\sum_i P_i \tau_i^{2}}{\sum_i P_i}
- \Bigl(\tfrac{\sum_i P_i \tau_i}{\sum_i P_i}\Bigr)^{2}}\) \\
Angular Spread (AS) &
\(\mathrm{AS}
= \sqrt{-2\ln\Bigl|\tfrac{\sum_i P_i e^{j\phi_i}}{\sum_i P_i}\Bigr|}\,[^\circ]\) \\
Rician \(K\)-factor &
\(K = 10\log_{10}\Bigl(\tfrac{P_{\mathrm{LoS}}}{\sum_{i\neq\mathrm{LoS}} P_i}\Bigr)\) \\
\bottomrule
\end{tabular}
\begin{tablenotes}
\item \(P_i\): received power of MPC \(i\); \(\tau_i\): excess delay of MPC \(i\);
\item \(\phi_i\): azimuth of MPC \(i\); \(P_{\mathrm{LoS}}\): LoS path power.
\end{tablenotes}
\end{threeparttable}
\vspace{-5mm}
\end{table}

\vspace{-2mm}
\subsection{Overall Channel Modelling Accuracy}
We first evaluate snapshot-level delay–spread model accuracy on the \textit{conference room} test scenes described in Section~\ref{sec:implementation}. Each snapshot corresponds to a fixed transmitter–receiver pair, and ground-truth RMS delay spread is computed from measured MPCs. We compare three classes of methods: (i) \textbf{\oursystem}, evaluated in two settings—\emph{zero-shot}, where material properties are inferred solely from vision priors, and \emph{few-shot}, where 20\% RF samples are used to calibrate material embeddings via differentiable optimization; (ii) \textbf{deep learning baselines}, represented by PointNet~\cite{mi_measurement-based_2024} trained with 100\% and 20\% of the dataset; and (iii) \textbf{RF3DGS}~\cite{zhang_rf-3dgs_2025}. We adapt the render to output delays and power per direction. To enable comparison on delay–spread metrics, we modify its output head to predict a per-direction delay value. Fig.~\ref{fig:exp-sota} reports the median and 75th-percentile RMS delay-spread errors for all methods. The median error of RF-3DGS, PointNet (20\%), PointNet (100\%), our zero-shot model, and our 10\% fine-tuned model is 11.8 ns, 3.8 ns, 1.5 ns, 2.5 ns, and 1.8 ns, respectively.
\begin{figure}[t!]
    \centering
    \includegraphics[width=0.7\linewidth]{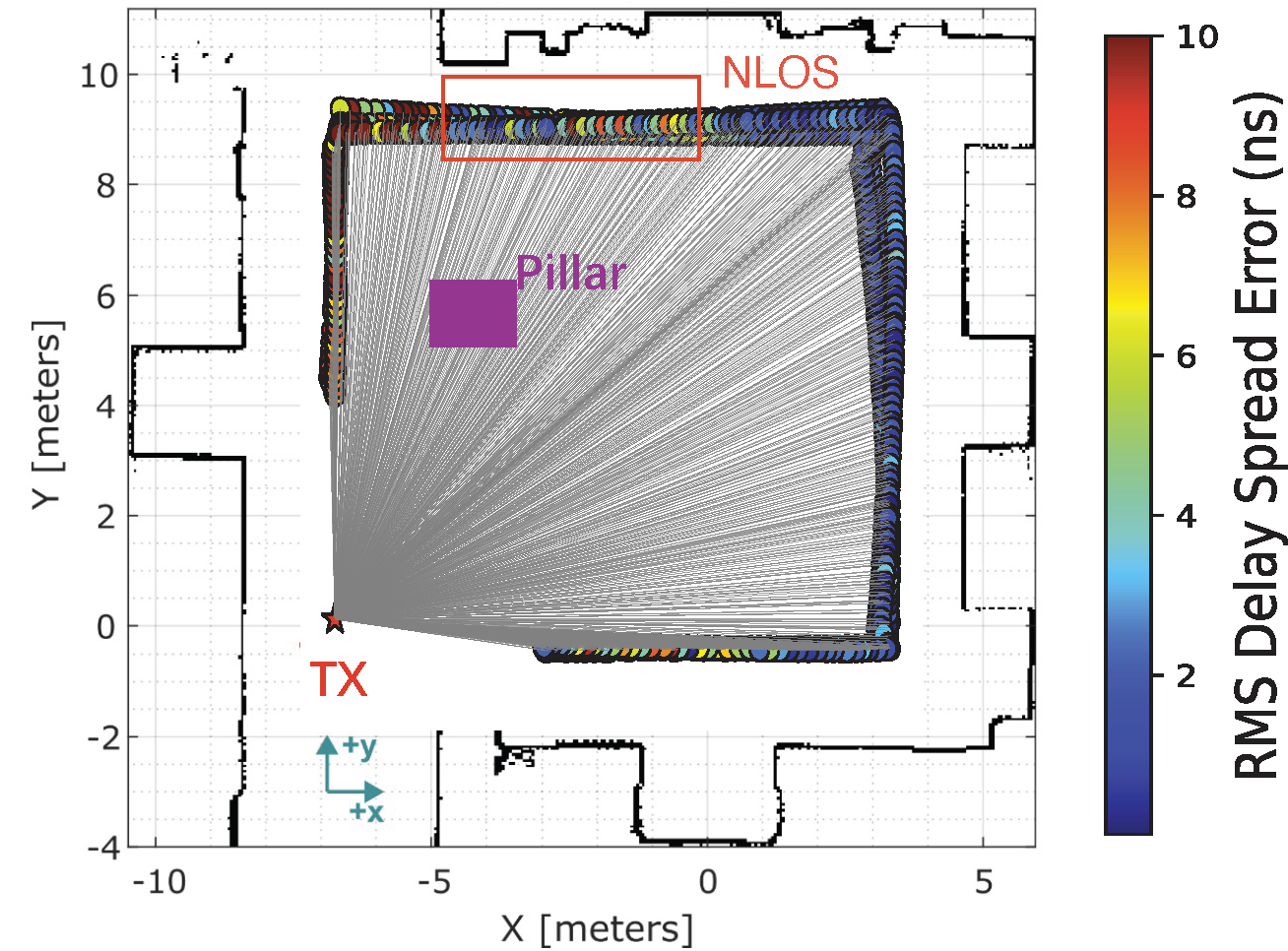}\vspace{-5mm}
    \caption{Spatial Distribution of RMS delay estimation}\vspace{-10mm}
    \label{fig:exp3-spatial}
\end{figure}

The comparison reveals several trends. (1) Even in the zero-shot, \oursystem a 34\% error reduction over PointNet, demonstrating that vision-derived material priors already impose meaningful physical structure for ray-based inference. (2), with only 10\% data, few-shot \oursystem substantially reduces errors by 59\%, narrowing the gap to the fully supervised PointNet model and highlighting the data efficiency enabled by differentiable calibration. (3), RF3DGS exhibits significantly higher errors. This is because it relies on an \emph{optical ray-marching} formulation that lacks a physical model for wireless multipath interactions—its radiance field can only fit per-direction intensity patterns, but cannot track the sequence of reflections and path lengths required to recover accurate multipath delay. Besides, as shown in Fig.\ref{fig:multipath-profile} and Fig.\ref{fig:exp-exp-multipath}, our measurements exhibit a sparse multipath profile, which poses a substantial challenge for optical-based radiance-field methods to coverage effectively. Overall, \oursystem offers accuracy competitive with deep learning while requiring orders of magnitude fewer channel samples.

Fig.~\ref{fig:exp3-spatial} further visualizes the spatial distribution of RMS delay spread model errors across the conference room. The heatmap shows that our system not only maintains accuracy in the direct LoS region near the transmitter but also generalizes well in non-LoS areas behind obstructions such as the central pillar.

\subsection{Snap-Shot Level Analysis}

We next analyze accuracy across the four canonical metrics in the \textit{conference room} test scene, as shown in Fig.~\ref{fig:metric-results}. We have the following observations:

\emph{(1) Delay spread}: As shown in Fig.~\ref{fig:rms}, Both our zero-shot and fine-tuning models successfully track the overall trend of the ground-truth RMS delay spread, while fine-tuning further suppresses overestimation in rich-scattering regions and yields closer alignment with the true multipath dispersion.

\emph{(2) K-factor}: the per-snapshot curves in Fig.~\ref{fig:k-factor} show that our predictions capture the ground-truth envelope well, with fine-tuning further reducing local deviations. The CDFs in Fig.~\ref{fig:cdf_k_delay} a quantify this gain: for K-factor, the median error drops from about $4$\,dB in the zero-shot case to $2$\,dB after fine-tuning, showing that calibration effectively corrects the LoS-to-NLoS power balance.

\emph{(3) Angular spread}. Fig.~\ref{fig:as} shows that zero-shot model track the diversity of multipath arrivals but occasionally exaggerate dispersion; fine-tuning sharpens the estimates and improves alignment with cluster structure.

\emph{(4) Path loss}. As shown in Fig.~\ref{fig:pl-1} and~\ref{fig:pl-2}, our model not only captures the short-term variations across samples but also recovers the long-term attenuation trend with distance, substantially outperforming the 3GPP InH-Office baseline, which consistently underestimates the slope. Together, these results demonstrate that \oursystem\ produces physically plausible zero-shot estimations across all metrics, while fine-tuning most benefits parameters sensitive to LoS dominance such as the K-factor.

\begin{figure}[t!]
    \centering
    \includegraphics[width=\linewidth]{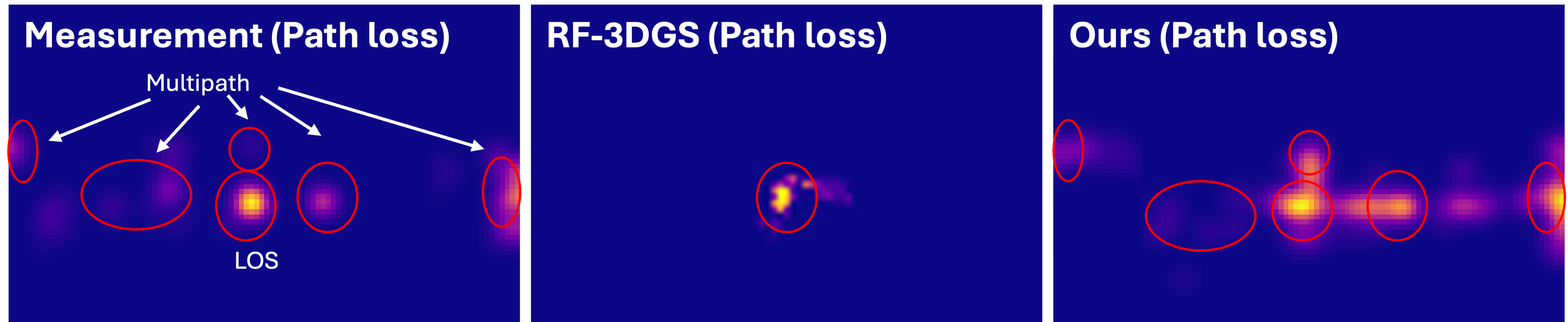}\vspace{-3mm}
    \caption{Multipath Profile Estimation. \textnormal{Power profile and delay profile comparison under zero shot estimation.}}\vspace{-5mm}
    \label{fig:multipath-ss}
\end{figure}

\begin{figure*}[t!]
    \centering
    \begin{minipage}{0.25\linewidth}
        \centering
        \includegraphics[width=\linewidth]{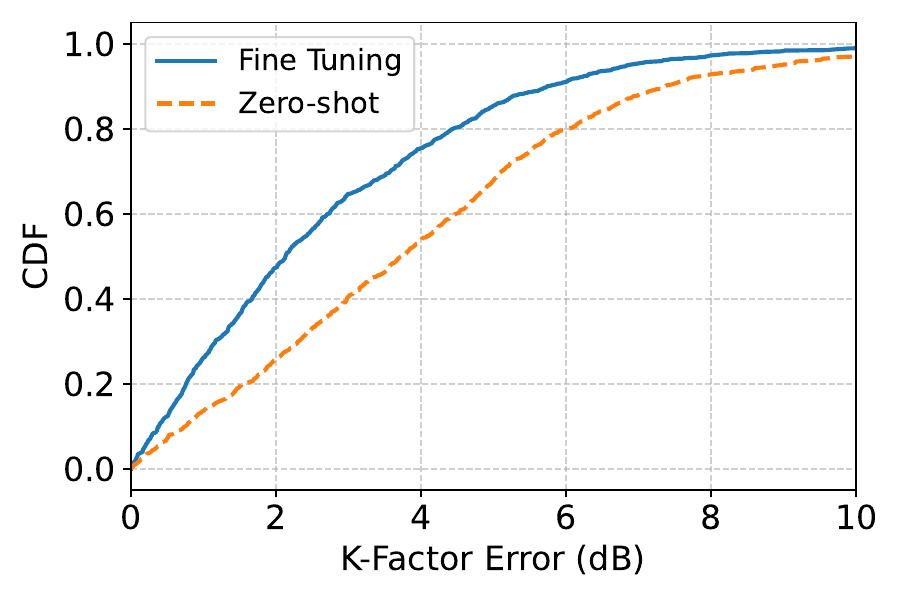}
        \vspace{-8mm}
        \caption{Fine Tuning.}\label{fig:cdf_k_delay}\vspace{-3mm}
    \end{minipage}%
    \begin{minipage}{0.25\linewidth}
        \centering
        \includegraphics[width=\linewidth]{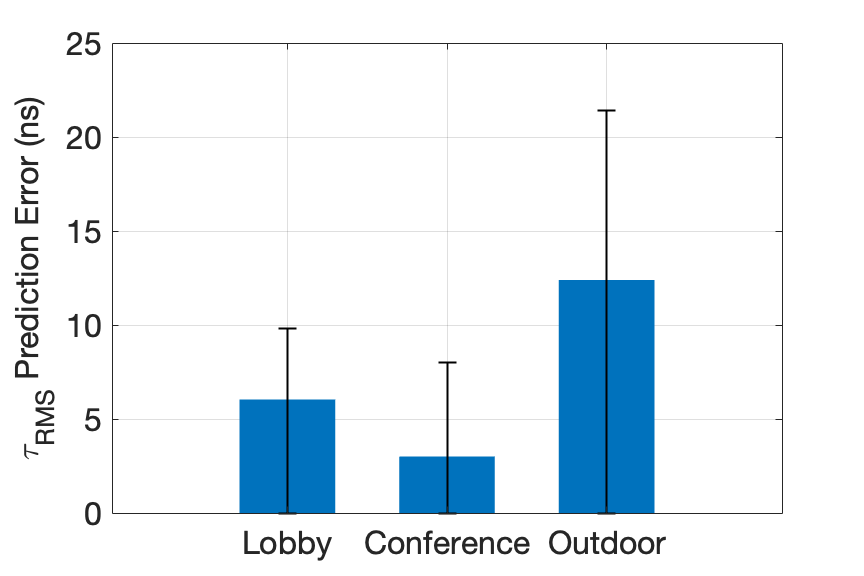}
        \vspace{-8mm}
        \caption{Across scenarios.}\label{fig:scenario_diversity}\vspace{-3mm}
    \end{minipage}%
    \begin{minipage}{0.25\linewidth}
        \centering
        \includegraphics[width=\linewidth]{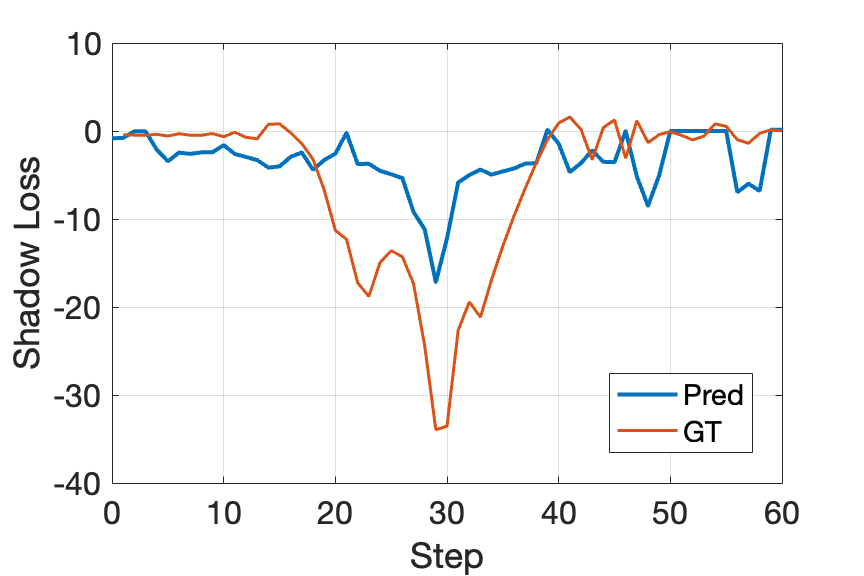}
        \vspace{-8mm}
        \caption{Mobility.}\vspace{-3mm}
        \label{fig:shadow-loss}
    \end{minipage}%
    \begin{minipage}{0.25\linewidth}
        \centering
        \includegraphics[width=\linewidth]{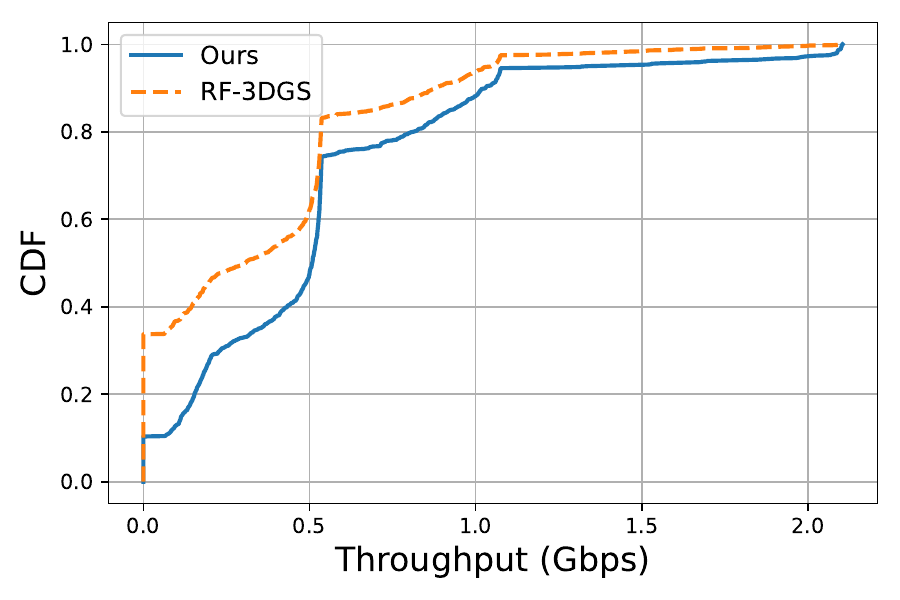}
        \vspace{-8mm}
        \caption{Case Study.}\vspace{-3mm}
        \label{fig:case}
    \end{minipage}
\end{figure*}

\subsection{Per-Path and Cluster-Level Analysis}
To obtain a fine-grained view of channel reconstruction, we examine per-path estimation in the angular domain. Fig.~\ref{fig:multipath-ss} plots the spatial spectrum of measurement, RF-3DGS and our estimation, color-coded by path loss. RF-3DGS captures the LoS path reasonably well but fails to reconstruct the remaining multipath structure. Because it relies purely on optical ray marching and is trained on sparse RF measurements, RF-3DGS cannot consistently track multipath components across viewpoints, leading to missing clusters and unstable predictions. \oursystem accurately recovers both the dominant LoS path and surrounding multipath clusters, with strong alignment in angle and power between estimation and measurement. The close correspondence indicates that our approach not only estimates aggregate metrics correctly but also resolves the underlying set and structure of multipath clusters, reducing false positives and missed paths in complex indoor scattering environments.

\subsection{Scenario Diversity}
To evaluate zero-shot cross-environment robustness, we perform a zero-shot test: no model is trained on the target scene. Each baseline operates only on its native inputs (RGB+LiDAR for \oursystem, geometry and standard ITU radio material parameters), and we measure the change in RMS delay-spread error when moving from indoor to outdoor deployments. Fig.~\ref{fig:scenario_diversity} shows the median absolute $\tau_{\mathrm{RMS}}$ estimation error for the Lobby, Conference-room, and Outdoor test scenes, with whiskers denoting the 90th percentile. Errors remain low indoors ($6.1$\,ns in the Lobby and $2.3$\,ns in the Conference-room) and increase moderately to $12$\,ns outdoors, reflecting the higher scattering complexity and less constrained geometry. Compared to differentiable ray tracing and PointNet-DL, which suffer substantially larger degradation across environments, \oursystem  maintains significantly lower error spread, demonstrating that vision-guided initialization enables strong generalization to unseen scenarios without retraining.

\vspace{-3mm}
\subsection{Evaluation under Mobility}
\begin{figure}[t!]
    \centering
    \includegraphics[width=0.85\linewidth]{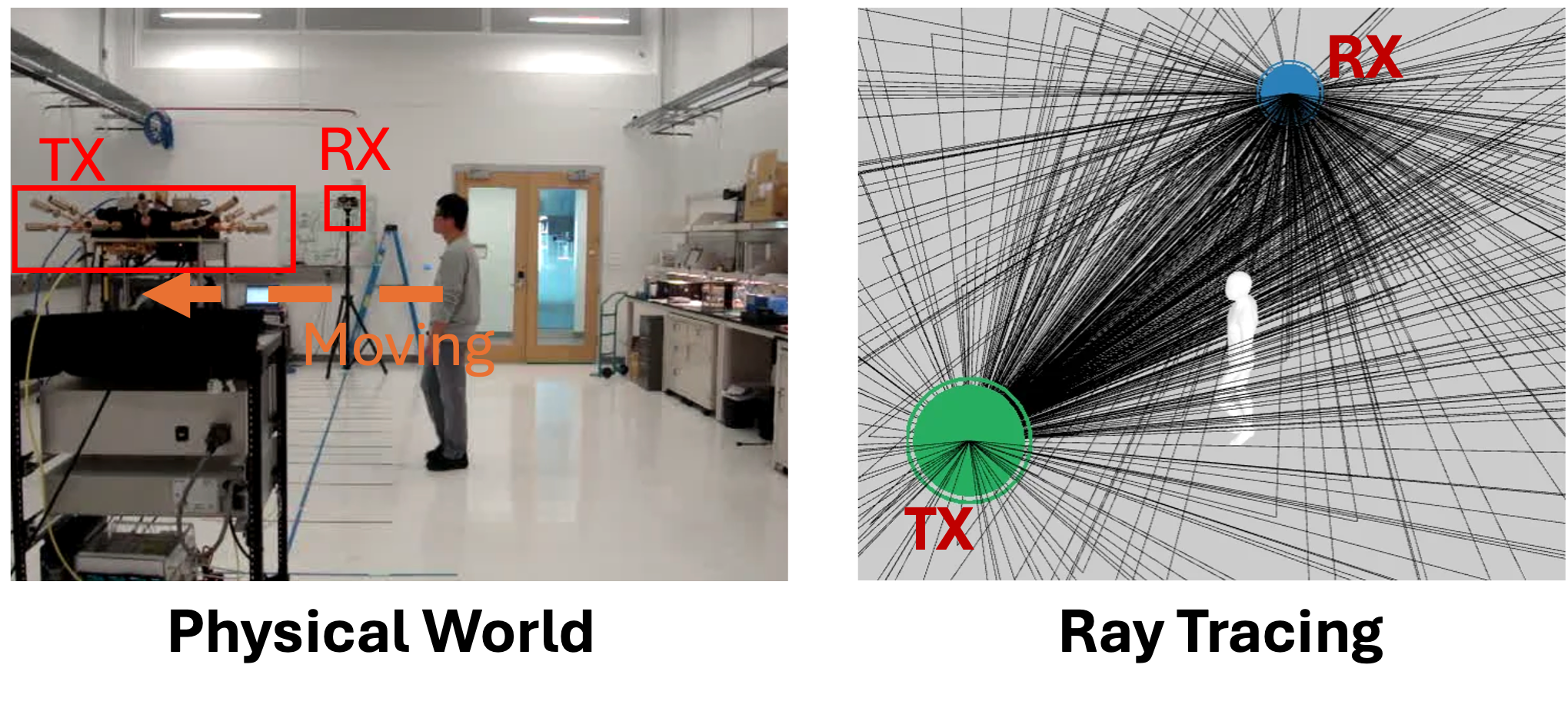}\vspace{-5mm}
    \caption{Dynamic Evlauation. \textnormal{(left) Lab measurement scene with a transmitter (TX) and receiver (RX) placed along a 4 m LoS while a human subject walks through the link. (right) Ray-tracing results.}}\vspace{-5mm}
    \label{fig:human-dynamic}
\end{figure}

To evaluate performance in dynamic environments, we consider the \textit{Dynamic-Human} scenario where a person walks at $1.2$\,m/s across the LoS between a transmitter (TX) and receiver (RX) separated by 4\,m (Fig.~\ref{fig:human-dynamic}). Every second, our system (i) captures a new camera frame, (ii) updates the local mesh to reflect the moving human, and (iii) re-traces propagation paths. We focus on shadow loss,
\[\small
SL_{\text{dB}} = 20 \log_{10} \left( \frac{|E_{\text{LoS}}|}{|E_{\text{Rx}}|} \right),
\]
where $E_{\text{LoS}}$ denotes the received field strength along the unobstructed LoS path and $E_{\text{Rx}}$ is the total received field during blockage. This ratio captures the relative attenuation caused by the human body crossing the link.

Fig.~\ref{fig:shadow-loss} plots the measured (red) and predicted (blue) shadow-loss traces over 60 walking steps. As the subject enters the link, both traces remain close to $0$\,dB until step $\sim$20, where the body begins to obstruct part of the LoS. A sharp deep fade of nearly $-35$\,dB occurs at step 30 when the subject fully blocks the link, after which the channel gradually recovers as the person continues walking. Our estimations not only reproduce the location and depth of this fade but also follow the fine-grained fluctuations before and after the blockage.

\vspace{-3mm}
\subsection{Runtime}
\label{sec:runtime}
A practical digital twin must support fast per-link inference. Once the geometry and material priors are constructed, the dominant runtime cost in \oursystem is a single forward pass of the ray tracer. Averaged over 500 transmitter–receiver pairs, the median end-to-end inference time is 0.33 seconds per link using the incremental ray tracing tricks in~\cite{an_radiotwin_2025}, comparable to neural ray tracing NeRF$^2$ (0.35s). This sub-second latency enables near–real-time channel estimation in quasi-static or slowly varying environments. When a local scene update occurs (e.g., a moving human), only the affected region is re-traced, still requiring sub-second latency. Thus, \oursystem supports rapid channel updates without retraining or rebuilding the digital twin.

\vspace{-3mm}
\subsection{Case Study: mmWave Beam Selection}
\label{sec:case}

To show how our vision-guided radio digital twin translates into real world gains, we conduct a case study following the beam selection procedure of SpaceBeam~\cite{woodford_spacebeam_2021}. For each Tx--Rx snapshot, both the naive geometry-only ray tracer and our system generate a set of MPCs (AoD, AoA, delay, and per-path path loss). We then select the best beam over all estimation: for each candidate beam, we compute the effective channel gain, derive the received power and SNR using the standard 2.16\,GHz noise model and receiver noise figure, and map SNR to an 802.11ad MCS index using SNR thresholds. The resulting PHY rate is further scaled by MAC-layer efficiency to obtain end-to-end throughput. This pipeline is identical for both ray tracers; the only difference is the underlying propagation model that produced the MPCs. As shown in Fig.~\ref{fig:case}, our method consistently achieves higher throughput across the scene. RF-3DGS frequently underestimates multipath power and mispredicts the dominant beam direction, particularly in NLoS settings. In contrast, our approach enables more accurate beam selection and stronger effective channel gains by preserving the underlying multipath structure. Overall, our system improves the median throughput by 30\%, confirming that accurate multipath reconstruction boosts mmWave performance.
\vspace{-3mm}
\section{Related Work}
\label{sec:related_works}

We review two major lines of research relevant to data-driven wireless channel modeling.
\vspace{-3mm}
\subsection{Channel Modeling and Prediction}

Prior research on channel prediction and ray tracing can be grouped into three main threads: \textbf{(1) End-to-End Deep Learning for Channel Prediction.} Early work~\cite{lee_pmnet_2023, narayanan_lumos5g_2020, mi_measurement-based_2024} employs classic neural networks to directly regress channel responses from raw inputs. While these models achieve high accuracy within their training domains, they typically overfit to environment-specific datasets and offer limited interpretability, as they do not explicitly disentangle the roles of scene geometry or material properties in radio propagation. \textbf{(2) Differentiable Ray Tracing with Physics Priors.} Recent advances (e.g., NeRF$^2$~\cite{zhao_nerf2_2023}, WiNeRT~\cite{orekondy_winert_2022}, NeWRF~\cite{lu_newrf_2024}, RF-3DGS~\cite{zhang_rf-3dgs_2025}) have introduced differentiable ray-tracing frameworks that embed physical priors into neural architectures, augmenting learned predictors with explicit geometric and electromagnetic constraints. While such methods improve physical fidelity and support gradient-based optimization, they are based on optical ray marching and do not have accurate radio physical model. They still require substantial per-environment training and generally achieve only limited transferability, even when incorporating vision-based cues. \textbf{(3) Vision-Based Material Segmentation for Ray Tracing.} Systems like mmSV~\cite{kamari_mmsv_2023} and the NIST Digital Twin~\cite{gentile_context-aware_2024} leverage computer vision to assign surface material labels for classical or hybrid ray-tracing simulation. However, these pipelines depend on narrow, closed-set material classifiers and produce coarse-grained labels as shown in the Fig.~\ref{fig:sys-segmentation}, lacking integration with differentiable tracers. As a result, they are unable to support end-to-end optimization, zero-shot adaptation, or principled uncertainty quantification. To reduce the overhead of channel measurement, recent work has injected visual cues into the modelling loop.  Systems such as mmSV~\cite{kamari_mmsv_2023} and NIST’s context-aware twin~\cite{gentile_context-aware_2024} segment materials in individual camera views, project the masks onto LiDAR point clouds, and then ray-trace the annotated mesh, demonstrating that even coarse imagery can reduce sounding budgets.  Yet their view-by-view processing propagates 2-D errors into 3-D, and their fixed, limited vocabularies overlook subtle but radio-relevant details such as surface roughness or gloss.

In summary, while these research threads have advanced the state-of-the-art in data efficiency, physical realism, and material-aware simulation, each suffers from critical limitations: poor generalization, a need for extensive environment-specific retraining, or reliance on coarse, out-of-domain material annotations. Our proposed approach addresses these gaps by decoupling geometry, material properties, and propagation physics. Furthermore, our appropach leverages large pre-trained vision–language models for open-vocabulary, zero-shot semantic material segmentation—enabling fine-grained, explainable, and transferable channel prediction in arbitrary, dynamic scenes.
\vspace{-3mm}
\subsection{Vision-Language Models}

Recent progress in large vision–language models (VLMs) such as Contrastive Language–Image Pre-training (CLIP)~\cite{openai_hello_2024} and Generative Pre-trained Transformer (GPT)-4o~\cite{openai_hello_2024}  has enabled powerful open-vocabulary scene understanding and multimodal reasoning~\cite{kerr_lerf_2023,engelmann_opennerf_2024,peng_openscene_2023,qiu_feature_2024}, but these capabilities have been largely confined to traditional vision or language applications, leaving the integration of VLMs into wireless ray-tracing for channel modeling  largely unexplored.
Pioneer work OpenScene~\cite{peng_openscene_2023} and Segment3D~\cite{huang2024segment3d} utilizes VLM-driven segmentation to annotate 3D scene reconstructions represented as sampled point clouds, but its performance is inherently constrained by mesh resolution, limiting its ability to capture fine-scale structures and materials—crucial for accurate electromagnetic modeling. To the best of our knowledge, our work is the first to integrate large VLMs with differentiable ray tracing for wireless propagation modeling, treating radio as a novel modality within a vision–language foundation. By bridging these domains, our system enables VLMs to inform physics-driven simulation, delivering interpretable, zero-shot channel prediction in complex, dynamic environments.

\vspace{-3mm}
\section{Conclusion}
\label{sec:concl}

In this work, we envision a new generation of wireless digital twins that are data-efficient, explainable, and adaptive to real-world dynamics. We introduce \oursystem, a zero-shot, vision-guided mmWave channel prediction system that combines pretrained vision–language models and physics-informed ray tracing. By translating rich visual semantics into radio-relevant parameters, our method delivers accurate, real-time channel prediction in arbitrary environments without the need for RF measurements or retraining. This approach paves the way for scalable, intelligent wireless systems and sets a foundation for future integration of vision and radio sensing.

\section{Acknowledgment}
This work is supported by National Science Foundation under Grant Nos. CNS-2433914, and CNS-2433915. The Princeton Advanced Wireless Systems research group gratefully acknowledges a gift from InterDigital Corporation.
\vspace{-5mm}

\bibliographystyle{ACM-Reference-Format}
\bibliography{references,sample-base}

\appendix
\section{Appendix}

This appendix provides additional implementation details, intermediate results, and visualizations that complement the main paper. We present (i) the full point-cloud processing pipeline used to obtain meshes for ray tracing, (ii) geometry-reconstruction accuracy benchmarks, (iii) our outdoor-scene reconstruction method using Google Earth Studio, (iv) the LLM-based material proposal process, and (v) the measured mmWave multipath profiles used as supervisory signals. These extended materials provide deeper insight into the engineering workflow behind \oursystem and validate the robustness of each module beyond what can be included in the main sections.

\subsection{Point Cloud Processing Pipeline}
\label{appendix:pcd}

To construct an accurate and watertight 3D mesh for ray tracing, we process NeRF-derived point clouds through a multi-stage pipeline. We first query the NeRF model along a sequence of virtual camera poses; at each viewpoint, the model predicts depth for every pixel, which is then back-projected into 3D to form a dense but noisy point cloud. As raw NeRF depth renders often contain floating points, holes, and view-dependent artifacts, we apply a filtering pipeline using MeshLab~\cite{meshlab}, including statistical outlier removal and voxel-grid downsampling.

\begin{figure}[h!]
    \centering
    \includegraphics[width=\linewidth]{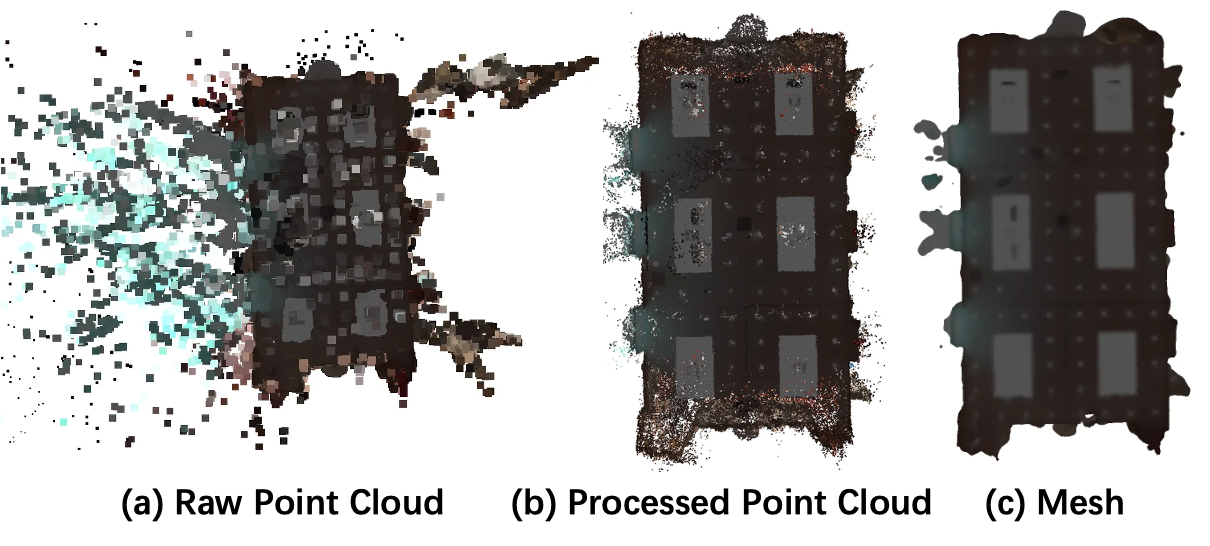}\vspace{-5mm}
    \caption{Progressive refinement of NeRF-derived point clouds. \textnormal{(1) raw point cloud generated by querying NeRF with virtual camera depth rendering, (2) filtered and down-sampled point cloud, and (3) final mesh obtained through Possion surface reconstruction.}
}\label{fig:point-cloud}\vspace{-5mm}
\end{figure}

The cleaned point cloud is then converted into a watertight mesh via Poisson surface reconstruction, followed by Laplacian smoothing to improve surface continuity. This final mesh serves as the geometric foundation for precise ray-tracing simulations in our digital twin. Fig.~\ref{fig:point-cloud} illustrates this progressive refinement process: (a) raw NeRF point cloud, (b) filtered point cloud, and (c) final mesh suitable for high-fidelity propagation modeling.

\subsection{Geometry Error Analysis}~\label{appx:geo-error}
A reliable radio–digital twin hinges on two pillars: (i) geometric
accuracy, which governs time delay, and (ii) material
identification accuracy, which controls reflection, transmission,
and attenuation coefficients. We therefore report both metrics separately.

Fig.~\ref{fig:exp0-ply-error} (left) visualizes the per‐vertex Euclidean distance between our mesh and the LiDAR ground truth, while
Fig.~\ref{fig:exp0-ply-error} (right) plots the corresponding histogram. Overall, 81\% of vertices lie within \SI{7}{\centi\metre} of the LiDAR surface.

\textbf{Discussion:} Most large residuals appear at distant room corners and ceiling light fixtures—regions where the camera observes the surface from far or oblique angles, causing depth-estimation drift. These areas are far from the transceiver and contribute minimally to mmWave propagation; conversely, the primary reflectors (walls, large furniture, glass partitions) exhibit errors below \SI{4}{\centi\metre}. According to prior analysis (e.g., RFCanvas~\cite{chen_rfcanvas_2024}, Fig.~16), \textit{{geometry deviations of this magnitude introduce less than 1~dB channel error}}, implying negligible impact on multipath prediction. Such corner drift can also be mitigated by repositioning cameras during data capture.

\begin{figure}[h!]
    \centering
    \begin{subfigure}{0.6\linewidth}
        \centering
        \includegraphics[width=\linewidth]{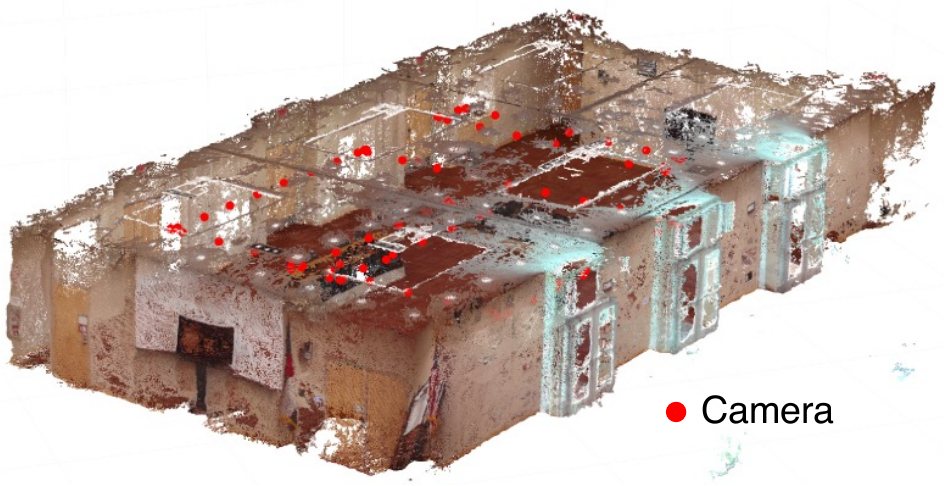}
        \caption{Camera trajectory}
        \label{fig:sub2}
    \end{subfigure}
    \begin{subfigure}{0.39\linewidth}
        \centering
        \includegraphics[width=\linewidth]{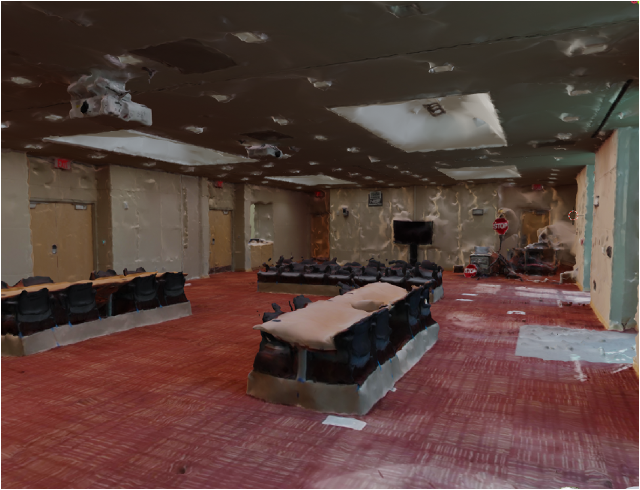}
        \caption{Reconstructed scene}
        \label{fig:sub1}
    \end{subfigure}
    \caption{Illustration of two reconstructed indoor scenes shown within a single column. \textnormal{(a) a second view emphasizing spatial layout and camera placement (b) A first-person reconstruction highlighting interior structures.}}\vspace{-5mm}
    \label{fig:twocol}
\end{figure}

\begin{figure}[h!]
    \centering
    \includegraphics[width=0.95\linewidth]{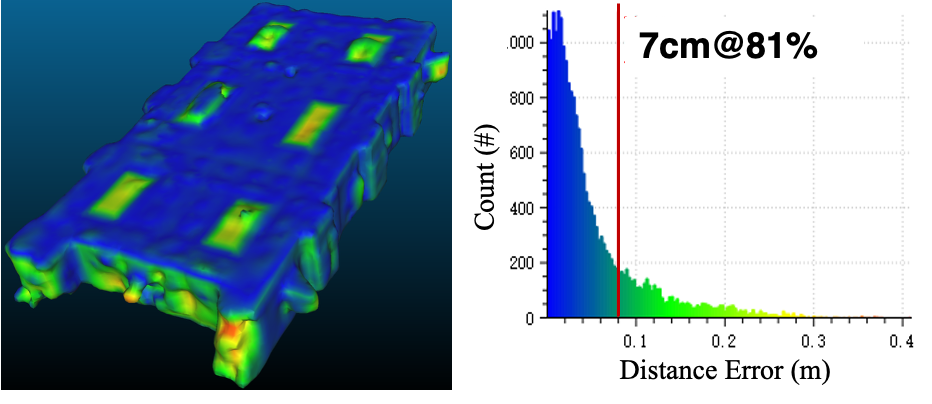}\vspace{-5mm}
    \caption{Vertex-wise geometry error of the mesh with respect to the LiDAR ground truth. \textnormal{left—spatial error map; right: mesh to mesh distance histogram.}}\vspace{-3mm}
    \label{fig:exp0-ply-error}
\end{figure}

\subsection{Outdoor environment reconstruction}
For large outdoor scenes, capturing dense multi-view images is difficult due to lighting variations, limited access, and the lack of stable camera poses. To address this, we \textit{virtually take photographs} in Google Earth Studio. We import the trajectory into Google Earth Studio and place a virtual camera. Google Earth Studio then renders a sequence of high-resolution frames together with accurate intrinsic and extrinsic parameters. These synthetically captured “virtual photos” provide consistent, well-posed multi-view supervision, enabling NeRF reconstruction and semantic feature extraction even in outdoor areas where physical image collection would be impractical.

\begin{figure}[h!]
    \centering
    \includegraphics[width=0.95\linewidth]{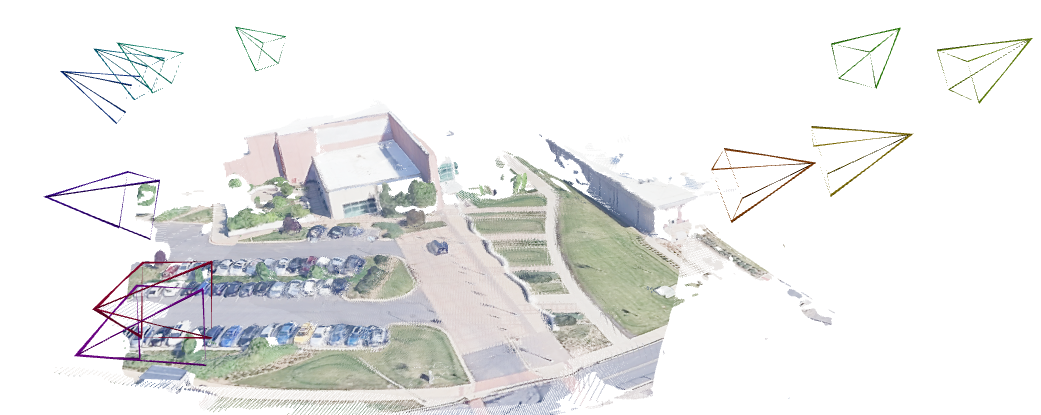}
    \caption{Outdoor dataset Reconstruction}
    \label{fig:outdoor}
\end{figure}

\subsection{LLM based Material Proposal}
\begin{figure}[h!]
    \centering
    \includegraphics[width=0.95\linewidth]{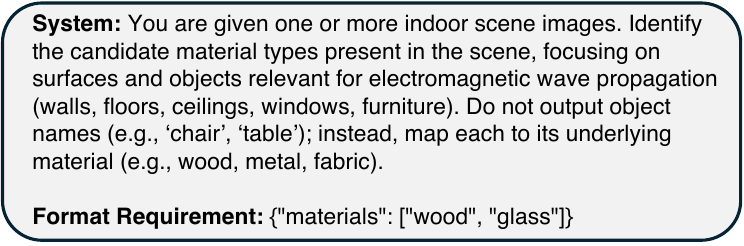}\vspace{-5mm}
    \caption{Material proposal prompt for GPT-4o.}\vspace{-5mm}
    \label{fig:sys-material-prompt}
\end{figure}

To obtain an initial set of plausible material categories for each scene, we query a large vision–language model (ChatGPT-4o) using the structured prompt shown in Fig.~\ref{fig:sys-material-prompt}. The model analyzes the input RGB images and returns a concise list of underlying material types (e.g., \texttt{"wood"}, \texttt{"glass"}, \texttt{"metal"}, \texttt{"plasterboard"}), avoiding object-level labels. These LLM-proposed materials form an open vocabulary from which we later identify point-wise material types via CLIP similarity matching. This step is crucial because it constrains the search space to realistic materials present in the scene while allowing generalization to unseen environments through the LLM's broad world knowledge.

\begin{figure}[h!]
    \centering
    \includegraphics[width=0.95\linewidth]{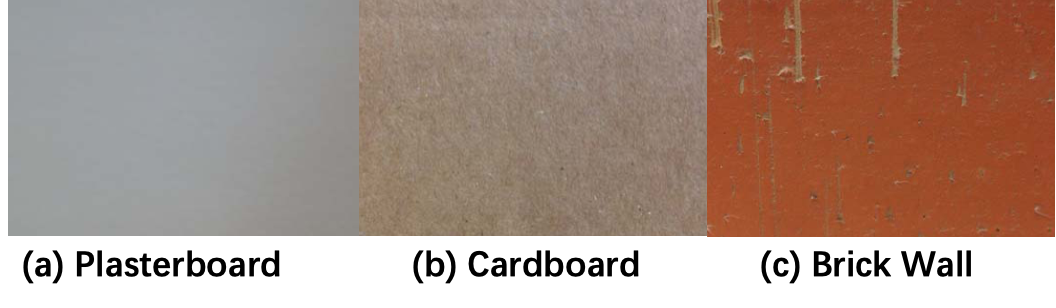}
    \vspace{-5mm}
    \caption{Appearance of materials with different roughness. \textnormal{As roughness increases, visual cues become more pronounced, enabling vision-based inference of surface roughness and corresponding scattering coefficients.}}\vspace{-5mm}
    \label{fig:scattering-material}
\end{figure}

\subsection{Multipath Profile Measurement}\label{appx:multipath-profile}
\begin{figure}[h!]
    \centering
    \includegraphics[width=0.95\linewidth]{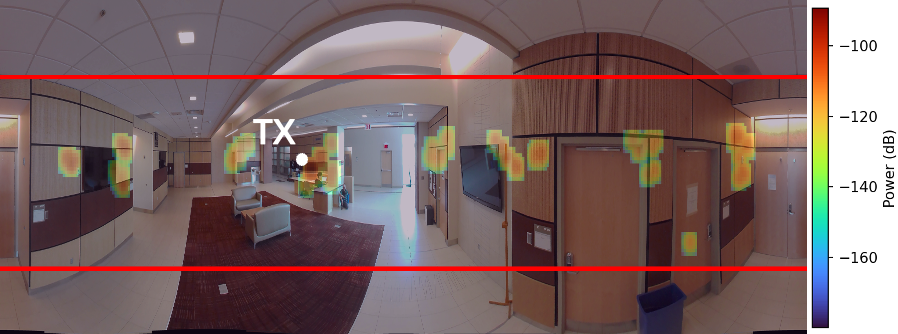}\vspace{-3mm}
    \caption{Measured Multipath Profile}\vspace{-3mm}
    \label{fig:multipath-profile}
\end{figure}

The ground-truth channel data was collected using the mmWave array channel sounder described in~\cite{gentile_context-aware_2024}. At each measurement snapshot, the sounder resolves individual multipath components (MPCs) by jointly estimating their delay, azimuth/elevation angles, and path-loss, yielding a high-resolution directional multipath profile (Fig.~\ref{fig:multipath-profile}). These MPCs provide the supervision for evaluating our channel-prediction accuracy and serve as the calibration targets used during differentiable ray-tracing refinement.
\clearpage
\end{document}